\documentclass[preprints,article,accept,moreauthors,pdflatex]{Definitions/mdpi}
\usepackage{subcaption}
\usepackage{amsmath}
\usepackage{array}
\newcolumntype{P}[1]{>{\centering\arraybackslash}p{#1}}

\firstpage{1} 
\pubvolume{xx}
\issuenum{1}
\articlenumber{5}
\pubyear{2019}
\copyrightyear{2019}
\history{Received: date; Accepted: date; Published: date}

\Title{FuseVis: Interpreting neural networks for image fusion using per-pixel saliency visualization}

\Author{Nishant Kumar $^{1*}$, Stefan Gumhold $^{1}$}

\AuthorNames{Nishant Kumar, Stefan Gumhold}

\address{%
$^{1}$ \quad Chair of Computer Graphics and Visualisation, Faculty of Computer Science, Technische Universit\"{a}t Dresden, N\"{o}thnitzer Stra{\ss}e 46, 01187 Dresden, Germany}

\corres{Correspondence: nishant.kumar@tu-dresden.de; Tel.: +49-351-463-38354}

\abstract{Image fusion helps in merging two or more images to construct a more informative single fused image. Recently, unsupervised learning based convolutional neural networks (CNN) have been utilized for different types of image fusion tasks such as medical image fusion, infrared-visible image fusion for autonomous driving as well as multi-focus and multi-exposure image fusion for satellite imagery. However, it is challenging to analyze the reliability of these CNNs for the image fusion tasks since no groundtruth is available. This led to the use of a wide variety of model architectures and optimization functions yielding quite different fusion results. Additionally, due to the highly opaque nature of such neural networks, it is difficult to explain the internal mechanics behind its fusion results. To overcome these challenges, we present a novel real-time visualization tool, named $\it{FuseVis}$, with which the end-user can compute per-pixel saliency maps that examine the influence of the input image pixels on each pixel of the fused image. We trained several image fusion based CNNs on medical image pairs and then using our \textit{FuseVis} tool, we performed case studies on a specific clinical application by interpreting the saliency maps from each of the fusion methods. We specifically visualized the relative influence of each input image on the predictions of the fused image and showed that some of the evaluated image fusion methods are better suited for the specific clinical application. To the best of our knowledge, currently, there is no approach for visual analysis of neural networks for image fusion. Therefore, this work opens up a new research direction to improve the interpretability of deep fusion networks. The \textit{FuseVis} tool can also be adapted in other deep neural network based image processing applications to make them interpretable.}


\begin{document}


\section{Introduction}

The recent development of state-of-the-art imaging modalities has revolutionized the way we perform our everyday activities. For example, in self-driving cars, infrared images from the respective camera sensors positioned at the vehicle help to detect obstacles such as pedestrians at night. Remote sensing satellites, on the other hand, acquires multi-spectral and multi-resolution images that are needed for object detection and recognition from high altitudes. In medical diagnosis and treatment, Magnetic Resonance Images (MRI), provides a detailed view of the internal structures of the human brain such as white and gray matter whereas Positron Emission Tomography (PET) and single photon emission computed tomography (SPECT) images provide functional information like glucose metabolism and extent of cerebral blood flow (CBF) or perfusion activity in the specific regions of the brain. However, it is challenging to analyze such complementary information provided by these image modalities individually. Multi-modal image fusion based image processing technique solves this problem by combining two or more pre-registered images from single or multiple imaging modalities into a fused feature space. Ideally, a fusion algorithm should have low computational overload and it should also preserve input features for usage in the aforementioned real-world applications.

In the early years of image fusion research, most of the methods had been proposed in a three-step approach. Firstly, the input images were converted into several feature maps using an appropriate image transformation method such as pyramid decomposition, wavelet transformation, and sparse representation. Then, the coefficients of the multimodal feature maps were combined using a pre-defined fusion strategy to get fused feature maps. Finally, the fused image was reconstructed using the inverse transformation applied to the fused feature maps. However, for an ideal fusion model design, these methods mainly focused on enhancing the transformation and fusion strategies to pursue good perceptual results by defining some intricate design rules. Recently, several unsupervised machine learning based CNNs have been proposed to achieve real time image fusion results. However, these image fusion based neural networks lacks the trust of the end-users since there are no tools to analyze the quality of the fused images as these networks act as blackboxes.  

To overcome this challenge, the neural network-based image fusion methods evaluate the quality of the results using performance metrics such as Structural Similarity Index (SSIM) \cite{ref9} that compares the perceptual similarity between the input images and the fused image. However, the shortcoming of such evaluation is that these metrics do not visualize the influence of input image features on the features of the fused image as it does not consider the underlying heuristics of the hidden network layers thereby not analyzing the internal mechanics of these networks. Therefore, image fusion using neural networks that have been evaluated on these performance metrics have low interpretability.   

Recently, several visualization techniques have been proposed that broadly try to interpret the neural network decisions. For example, the gradient based visualization methods backpropagate the gradients through the hidden layers of the neural network to interpret the per-pixel influence of the input image on the output predictions. However, such methods are specifically suited for problems such as image classification where the aim is to visually explain the class decision made by the neural network using jacobian based saliency map for the class score with respect to the input image. On the contrary, the primary goal for interpreting per-pixel decisions in a fusion problem is to compute the jacobian based saliency map for each pixel of the fused image with respect to each pixel of the input image. However, such analysis leads to very high computational overload since the number of backpropagation iterations is equivalent to the number of fused pixels. To circumvent this challenge, the deep learning-based software frameworks such as PyTorch and TensorFlow implicitly aggregate the per-pixel gradients from the input tensor elements to compute a single gradient value for each of the output tensor elements. In image fusion, this is not helpful since the gradient information related to how each pixel in the input image influences the machine decision is lost. Hence, the per-pixel saliency visualization by computing gradients of a neural network based fusion model remains untapped in popular literature. 

Therefore, in this paper, we present a novel per-pixel saliency based visualization approach with real time capabilities that will guide in the interpretability of image fusion based neural networks. The major novelties of this work are summarised below:

\begin{itemize}
  \item We trained several state-of-the-art fusion based unsupervised CNNs under the same learning objective. Considering the importance of understanding the fusion blackboxes in life sensitive domain such as medical imaging, we focused on interpreting the trained neural networks specifically for MRI-PET medical image fusion. 
  \item We performed fast computation of per-pixel jacobian based saliency maps for the fused image with respect to input image pairs. To the best of our knowledge, it is a first-of-its kind technique to visualize fusion networks by considering the backpropagation heuristics that helps it to be more transparent in a real-time setup.
  \item We constructed guidance images for each input modality by using gradients of the fused pixel with respect to the input pixel at the corresponding location in the input image. We also interpreted the gradient values in each of the guidance images with the grayscale intensities of the fused image by combining the images in the color channels of an RGB image.
  \item We computed scatter plots between the gradients of the guidance images which provides a visual overview of the correlation between the influence of each of the input modalities. For example, a positive correlation will show that the input modalities influence the fused image equally. 
  \item We developed an interactive Graphical User Interface (GUI) named $\it{FuseVis}$, that combines all the visual interpretation tools in an efficient way. The FuseVis tool allows to compute saliency maps in real-time on the mouse over at the pixel pointed to by the mouse pointer. Our code is available at \href{https://github.com/nish03/FuseVis}{https://github.com/nish03/FuseVis}.
  \item Finally, we performed clinical case studies on MRI-PET image pairs using our FuseVis tool and visually interpreted the fusion results obtained from several different neural networks. We showed the usefulness of FuseVis in identifying the capability of the evaluated neural networks to solve clinically relevant problems.
  
  \end{itemize}
  
 Section 2 provides a detailed literature review of the existing works relevant to the problem of image fusion and visualization of neural networks. In Section 3, we explain the visual analysis goals, the proposed visualization concepts, and our FuseVis tool layout in detail. In Section 4, we provide the experimental details of the training setup, the hardware used, the architecture of the neural networks for image fusion, loss function used, and the hyperparameter studies conducted. In Section 5, we present a clinical application of MRI-PET medical image fusion and discuss key visualization requirements for case studies in this field. In Section 6, we interpret the visualization results obtained from each of the fusion based neural networks using our FuseVis tool. In the same section, we also show the frame rates achieved during the mouseover operation that supports the real-time capabilities of our tool. Finally, in Section 7, we summarise the major contributions of this work and discuss its applicability in the field of real-time visualization of neural networks in general.


\section{Related Work}
In this section, we provide an overview of the literature relevant to this paper. For this, we divide this section into five categories. The first sub-section summarises the classical image fusion approaches that involve non-machine learning based techniques proposed in the recent past along with some relevant review papers related to image fusion problem. The second sub-section gives an overview of the methods that applied the classical image fusion approach along with pre-trained deep neural networks to attain the fusion results. The next sub-section discusses recent methods in the field of unsupervised end-to-end deep learning based image fusion specifically in the area of multi-focus and multimodal image fusion. The fourth sub-section gives a detailed overview of the recent state-of-the-art visualization techniques that have been proposed for interpreting and explaining the black-box neural networks in general. Finally, we end the related work discussion with current literature on the fast computation of gradients in a neural network setup.
 
\subsection{Classical Image fusion approaches}
Many review papers in the past such as \cite{ref1, ref2, ref10} provided an overview of different methods for multimodal and multi-focus image fusion. It is quite pertinent from these papers that the classical method for the fusion of multimodal images involves activity level measurement and pre-defined fusion rules. Activity level measurement aims to transform the input images to obtain salient features by either performing coefficient based activity measurement \cite{ref4}, window-based activity measurement \cite{ref36}, or region based activity measurement \cite{ref15}. The transformation strategies were mainly categorized into domains namely multi-scale decomposition, sparse representation, and hybrid transformation among others. This was followed by pre-defined fusion rules to output a fused image. Subsequently, the quality of the fused image was evaluated by using performance metrics that involve human visual perception and some objective statistical assessments. 

Multi-scale decomposition (MSD) \cite{ref4, ref11, ref12, ref13, ref18, ref14, ref15, ref16, ref17, ref19, ref20, ref21, ref22, ref23, ref24, ref25} separates the source image into low and high frequency sub-bands in order to separately analyze the base and detail level features. Some of the popular transformation strategies under multi-scale image decomposition has been pyramids \cite{ref4, ref11}, discrete wavelets \cite{ref12,ref13,ref14}, dual-tree complex wavelets \cite{ref15}, contourlets \cite{ref16} and shearlets \cite{ref17, ref18}. Multi-scale geometric analysis based MSD methods such as non-subsampled contourlet (NSCT) \cite{ref19, ref20} and non-subsampled shearlet transform (NSST) \cite{ref21, ref22, ref23} have also been proposed due to their effectiveness with image representation. Edge preserving filters such as bilateral filters \cite{ref24} and guided filters \cite{ref25} have been popular in multi-scale decomposition based literature since they retain the salient features from the input images. Sparse representation (SR) based transformation strategies \cite{ref5, ref27, ref28, ref29, ref30, ref42a} do not decompose the original image into low and high frequency bands but instead assume that both frequency bands have similar sparse coefficients. Hybrid transformation strategies aim to define more than one transformation procedure such as curvelet-wavelet \cite{ref31}, combination of multi-scale decomposition and sparse representation based methods \cite{ref32, ref33, ref34}, Intensity-Hue-Saturation (IHS) and Principal Component Analysis (PCA) \cite{ref35} among others. 

Finally, the image fusion rules after the image transformation have traditionally been defined to combine the features obtained from multi-scale transform approaches into a single fused image and been a crucial factor that helps to provide state-of-the-art fusion performances \cite{ref10}. The three different components which together comprised a robust fusion rule in the past are coefficient grouping, coefficient combination as well as consistency verification. The coefficient grouping and combination strategies include choose-maximum rules \cite{ref37}, weighted-averaging \cite{ref4} and guided-filtering based weighted averaging \cite{ref25}. Subsequently, the consistency verification aimed to ensure that the neighborhood coefficients are fused using the same fusion rule by refining the calculated weight map based on some priors \cite{ref38}. At last, the fused image obtained from the classical methods were evaluated with several performance metrics \cite{ref39, ref40, ref41, ref42, ref9} since there is no gold standard for an ideal fused image. 

\subsection{Mixture of classical and deep learning based fusion approaches}
In recent years, there have been several works that use deep learning based fusion networks \cite{ref3} to further enhance the fusion image quality. However, these works focused on using pre-trained neural networks to extract high and low-level image features for activity level measurement while the fusion rules were still manually defined. For example, in \cite{ref43}, a CNN for medical image fusion was used to perform activity level measurement and based on these measurements, coefficient grouping and coefficient combination were performed manually to obtain the fused image. On the other hand, in \cite{ref44}, the infrared and visible input images were first separated into low and high-frequency features using a pyramid based multi-scale decomposition approach, and then the high-frequency features were fed into a pre-trained deep learning based framework to extract multi-channel high-frequency features. Finally, the low-frequency features were fused using weighted averaging while the high-frequency multi-channel features were fused using a pre-defined max-selection rule. 
In \cite{ref7}, low-resolution grayscale images were divided into high and low-frequency features using wavelet transformation, and the high-frequency features were fed into a trained neural network which provided the high-resolution version of the input. Then, the absolute-max strategy was applied to the high-resolution outputs while the simple averaging was done to fuse the low-frequency features. Finally, the inverse wavelet transformation was done to retrieve a high-resolution output image. \cite{ref8} on the other hand, used multi-scale convolutional neural networks to obtain multi-scale feature maps that are fused and then post-processed to obtain the decision maps which are segmentation results of the fused image using watershed transformation. This is done to evaluate the quality of the fused image from different image fusion algorithms.

\subsection{Unsupervised end-to-end deep learning based fusion approaches}
 All the fusion methods discussed up to now use the traditional approach of defining the transformation and fusion strategies while some use deep neural networks to extract feature maps from its hidden layers to perform the activity level measurement. Based on the feature maps generated, a final fusion rule is defined. The second observation with these methods is that the network is trained on image distributions which possess properties quite distinct from the inferred images. Though such usage of convolutional neural networks has provided good results in these approaches, it undermines the true ability of a standalone end-to-end learning algorithm capable of providing a final fused result by eliminating pre or post-processing steps such as traditionally used transformation and fusion strategies. Recently, there are end-to-end deep learning based image fusion methods that introduce an optimization strategy using a neural network that models all the transformation, fusion, and reconstruction strategies in its hidden network layers and learns a fused image in an unsupervised manner using loss functions such as SSIM \cite{ref9}. \cite{ref6, ref45, ref50} were some of the early works in the field of multi-focus image fusion which used end-to-end convolutional neural networks that jointly generated activity level measurements and fusion rules. However, these methods trained the network in a supervised environment by leveraging the training data with available groundtruth from image classification databases. \cite{ref61} on the other hand, proposed a simple encoder-decoder based architecture for directly mapping the source images with a single camera focus to get a multi-focus fused image while \cite{ref54, ref59} proposed a Conditional Generative adversarial network (CGAN) based approach for multi-focus image fusion. In the field of multimodal medical image fusion, \cite{ref46} proposed a novel end-to-end fusion network to fuse MRI and PET image pairs by extracting high and low-frequency features within the network layers with SSIM as the loss function. \cite{ref47} used a similar approach for fusing multi-exposure RGB images. However, they first converted the RGB images into YCbCr color space and used an end-to-end neural network to fuse only the luma component $Y$ of the input images while the Cb and Cr components were fused using weighted averaging strategy. Finally, the fused YCbCr color space was converted into the RGB fused color space. There have been several works in the field of visible and infrared image fusion as well that utilize unsupervised deep learning based fusion framework. \cite{ref48} trained an end-to-end convolutional neural network using visible and infrared pedestrian images from both day and night to attain robust fusion results. \cite{ref49} used a different approach to fuse infrared and visible images where they used only a single modality at once to optimize the common weights. Secondly, the feature extraction and feature reconstruction layers involved the trainable weights while the fusion layer did not contain any trainable parameters. \cite{ref51} went a step further and defined a trainable fusion layer. \cite{ref53, ref57, ref60} applied the GAN approach for fusing infrared and visible images using end-to-end neural networks. \cite{ref55} was another such GAN based approach but they used a single modality and fused multi-resolution input images. Finally, there were some works \cite{ref56} which used a novel densely connected network using unsupervised learning strategy to fuse source images for multiple fusion tasks whereas \cite{ref58} proposed a new GAN based method that combined the identity of one input image and the shape of another input image.

\subsection{Visualization techniques}
Recently, there have been several methods that interpreted the decisions made by neural networks by visualizing the model predictions. \cite{ref62} proposed a method for visualization of classification models by computing the gradients of the class predictions with respect to the input images. \cite{ref63} on the other hand, computed the gradients for activation neurons in the hidden layers of the network with respect to the input image by visualizing the specific regions of the image that activates a given neuron the most. \cite{ref64} extended this work and generalized the gradient-based visual explanations proposed in \cite{ref63} for different types of neural network architectures. \cite{ref65} proposed an alternative visualization method where they used a relevance score approach that visualized the contributions of each pixel of the input image to the classifier predictions. \cite{ref66} presented a method inherently similar to \cite{ref65} but they also visualized the output prediction by calculating the contributions of all neurons in the network to every feature of the input image. 

Perturbation based visualization algorithms, however, do not need backpropagation heuristics and is a model agnostic approach.
\cite{ref67} was one of the first perturbation based approaches that tried to implement occlusion in the input image and examined the output of the classification based neural networks. The results showed that the model was not localizing the objects in the image as the probability of the correct class reduced significantly when the object was occluded. \cite{ref68} was another model agnostic method where the perturbations were performed in the input image around its neighborhood and the behavior of the model's predictions was later analyzed. Finally, the method weights the perturbed data with respect to its proximity to the original image and learns an interpretable model based on the old and the new predictions. \cite{ref70} interpreted the model predictions by changing the pixel intensities of the input images with some noise like blurring or occlusion and modeled the change in the prediction probability of the output image. \cite{ref69} showed that there are two important characteristics of a visualization method to properly explain the regions of the input that are important for the model's decisions. These characteristics are gradients and implementational invariance. To satisfy both the properties, the method integrated the gradients from multiple inputs due to which the method was able to satisfy the sensitivity and implementational invariance of a visualization method. Therefore, the method was able to visualize the prediction of a neural network with respect to its inputs and required just a few instances of gradient computations. \cite{ref71} proposed a visualization approach specifically for medical image fusion where they computed a heat map showcasing the mutual information between the source and the fused image. However, this method cannot interpret and explain the learning based fusion networks itself since there are no backpropagation heuristics involved. Although these perturbation based visualization approaches are generally applicable to a wide range of the problem, they still need several iterations to get the visualization results due to which they are not yet suitable for real-time deployment.

Crucially, none of the visualization methods can be generalized for fusion based deep neural networks since, in a fusion problem, the number of output predictions is equal to the number of output pixels. This has huge memory and time requirements due to the very high number of backpropagation heuristics involved. \cite{ref72} efficiently computed the gradients of the loss function by conducting the backpropagation heuristics for per-training example. However, the idea only works on a neural network for image classification that has a single class prediction, and therefore, it cannot be applied to an image fusion problem where there are a large number of predictions in the form of fused pixels. Therefore, an efficient method is needed to decrease the computational cost for the visualization of high dimensional output predictions provided by deep neural networks.

\section{Method}
In this section, we will present the key visual analysis goals of a neural network for image fusion. Next, we will describe the proposed visualization concepts along with their mathematical formulations that help in achieving these visual analysis goals. Finally, we will present an overview of our user interface, FuseVis, and describe its user interaction capabilities.

\subsection{Visual analysis goals}
A fused image provides important information related to the input image modalities in combined feature space by preserving features from both the input images. However, by visualizing only the fused images, it could be challenging to interpret the sensitivity of the fusion methods for the input images. It is especially difficult to interpret which of the fusion methods reproduce features from which input image in a better way and how sensitive it is to changes in pixel intensities of the input images. Hence, this analysis is important to analyze the stability of a fusion network to the pixel intensity changes in the specific regions of the input images. 

\begin{figure}[!htb]
\centering
\includegraphics[width=15 cm]{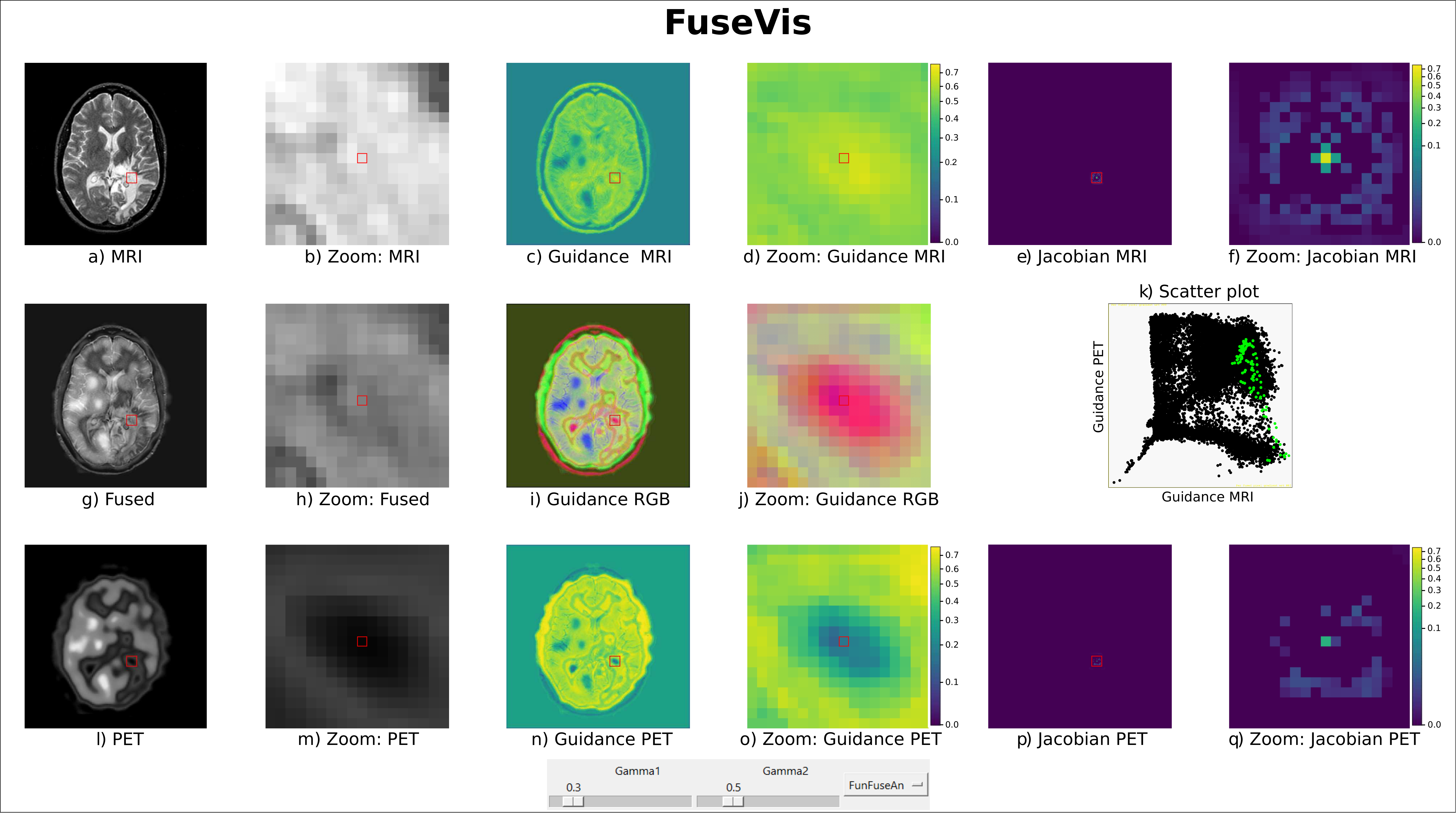}
\caption{The figure shows a layout of FuseVis's user interface.}
\label{fig:fig1}
\end{figure} 

\subsection{Visualization concepts}
We present several visualization strategies that help to understand the fusion results. The most important of those strategies is to compute gradient based saliency heat maps that indicate the relevance of each input pixel to a single pixel in the fused image. These heatmaps are particularly valuable since they allow for an easy and intuitive investigation of what the respective fusion model perceived to be important in the input image features. Therefore, gradient based saliency heatmaps are useful in interpreting fusion blackboxes and understanding the characteristics of image fusion based neural networks. Using these saliency based heat maps, we can also estimate whether a neural network for image fusion is robust to new examples. Therefore, to improve the interpretability of such image fusion based neural networks, we present several visualization techniques that help to understand the fusion results.


\subsubsection{Jacobian images} We define per-pixel saliency heat maps, named as $\textit{Jacobian images}$ which highlight the regions in the input images that most prominently influenced the per-pixel fusion results. Let us interpret a fusion operator as a mapping $f:\mathbb{R}^{2n} \rightarrow \mathbb{R}^{n}$ that maps two input images $\boldsymbol{x_1}$ and $\boldsymbol{x_2}$ to a fused image $\boldsymbol{y} = f(\boldsymbol{x_1},\boldsymbol{x_2})$, where all images have the same dimensions and contain $n$ pixels. Then, the saliency analysis is focused on a single pixel $i$ $\in$ $\{1, ..., n\}$ of the fused image that can be selected by the user interactively and is called $\textit{principle pixel}$, $y^{(i)}$, where we use a superscript enclosed in parenthesis to denote the pixel of an image. For each input image $\boldsymbol{x}$ $\in$ $\{\boldsymbol{x_1},\boldsymbol{x_2}\}$, we define the jacobian image of $y^{(i)}$ as the image of the partial derivatives with respect to the input image $\boldsymbol{x}$:


\begin{equation}
\label{eqn:label1}
\frac{\partial y^{(i)}}{\partial \boldsymbol{x}}   = 
\Bigg(
  \frac{\partial y^{(i)}}{\partial x^{(1)}}, 
    \frac{\partial y^{(i)}}{\partial x^{(2)}}, ...., \frac{\partial y^{(i)}}{\partial x^{(i)}}, 
    ....,
    \frac{\partial y^{(i)}}{\partial x^{(n)}}\Bigg)  
\end{equation}

 The jacobian computation involves backpropagating the gradients through the hidden layers of the neural network using chain rule-based automatic differentiation. A jacobian image visualizes the extent up to which each pixel element of the input image $\boldsymbol{x}$ influences the principle pixel $y^{(i)}$. Therefore, the jacobian images reveal the sensitivity of the principle pixel $y^{(i)}$ in the fused image to the changes in the pixel intensities of the input images. An example of a jacobian image is shown in Figure \ref{fig:fig1} e). It can be expected that the fused principle pixel $y^{(i)}$ is highly sensitive to changes in the grayscale intensities at  $x^{(i)}$. Additionally, the local neighborhood of the pixel element $x^{(i)}$ might also have a direct influence on the fused principle pixel $y^{(i)}$. The jacobian image based visualization concept also helps to compare the sensitivity of the fused principle pixel with respect to changes in the pixel intensities of each of the two input images as well as estimate which of the two input image has a greater neighborhood influence on the fused pixel. 

\subsubsection{Guidance images} Based on the observation that the pixel $x^{(i)}$ has by far the largest saliency value, we developed a visualization concept named $\textit{Guidance images}$ which only considers the gradients of the principle pixel $y^{(i)}$ with respect to the input pixel element $x^{(i)}$ located at the corresponding coordinate in the input image:


\begin{equation}
\label{eqn:label2}
\frac{\partial \boldsymbol{y}}{\partial \boldsymbol{x}}   = 
\Bigg(
  \frac{\partial y^{(1)}}{\partial x^{(1)}}, 
    \frac{\partial y^{(2)}}{\partial x^{(2)}}, ...., \frac{\partial y^{(i)}}{\partial x^{(i)}}, 
    ....,
    \frac{\partial y^{(n)}}{\partial x^{(n)}}\Bigg)  
\end{equation}

While jacobian images allow per-pixel interpretation of fusion outputs, guidance image aims to provide for each input image, a static overview of its influence on the fused image. An example of a guidance image is shown in Figure \ref{fig:fig1} c). The two static guidance images provide an overview of the jacobian images and allow us to compare the major gradient values on an absolute scale. This fosters comparison of the same pixel location in the two input modalities as well as different pixel locations in the same input image. The changes in the pixel intensities of the input image region with high gradients will significantly change the pixel intensities in the corresponding region of the fused image. On the contrary, it will require significant changes in the pixel intensities of the input image region with low gradients to have a similar effect on the fused features in the same region. Due to this, the guidance images reveal which of the
two input modalities have higher influence on the fusion result in specific regions of interest as well as where each input image has a large influence on the fusion result.

\subsubsection{Guidance RGB images}

During the interactive visual analysis with jacobian images and guidance images, a typical analysis task is to compare the values in the guidance images of the two modalities in order to find out which modality influences the fused image more. Furthermore, one typically wants to compare a guidance image with the fused image. To simplify these local comparisons, we provide a single Guidance RGB image that encodes the two normalized guidance images as well as the fused image in the three color channels red, green, and blue. We perform a max-min normalization scaling of the guidance images to re-scale the features with a distribution value between 0 and 1. For every guidance image, the minimum value of that image gets transformed into 0, and the maximum value gets transformed into 1. An example of a Guidance RGB image is shown in Figure \ref{fig:fig1} i). In red are the regions where the MRI gradient is high, the PET gradient is low and the fused image has relatively low pixel intensities. By increasing fused pixel intensities, the color changes from red to magenta. Similarly, green and cyan correspond to high PET gradients with low MRI gradient values. In blue regions, both gradient values are low and the fused pixels are bright. The yellow depict regions where both gradients are high. In this way, the Guidance RGB image allows us to find different gradient constellations fast and the selection of the principle pixel is also possible on the Guidance RGB image.

\subsubsection{Scatterplot}

As the Guidance RGB image gives a local overview of how the gradients of each modality behave in comparison to the fused image, a scatterplot can depict the correlation and more advanced statistical relationships between the gradient values of the guidance images. For example, a positive correlation between these gradients would mean that the input modalities equally influence the fused pixels while a negative correlation will show that an increase in the influence of one modality will lead to a decrease in the influence of the other modality.
For this statistical analysis of the gradient values in the guidance images, we show a scatterplot with MRI gradients along the x-axis and PET gradients along the y-axis with an example provided in Figure \ref{fig:fig1} k). The overall scatterplot in black shows a rather complicated relationship between the gradients. Therefore, we illustrated the points corresponding to pixels around the principle pixel in green. In this way, the local statistical relationships can be explored interactively.

\begin{figure}[!htb]
\centering
\includegraphics[width=15cm]{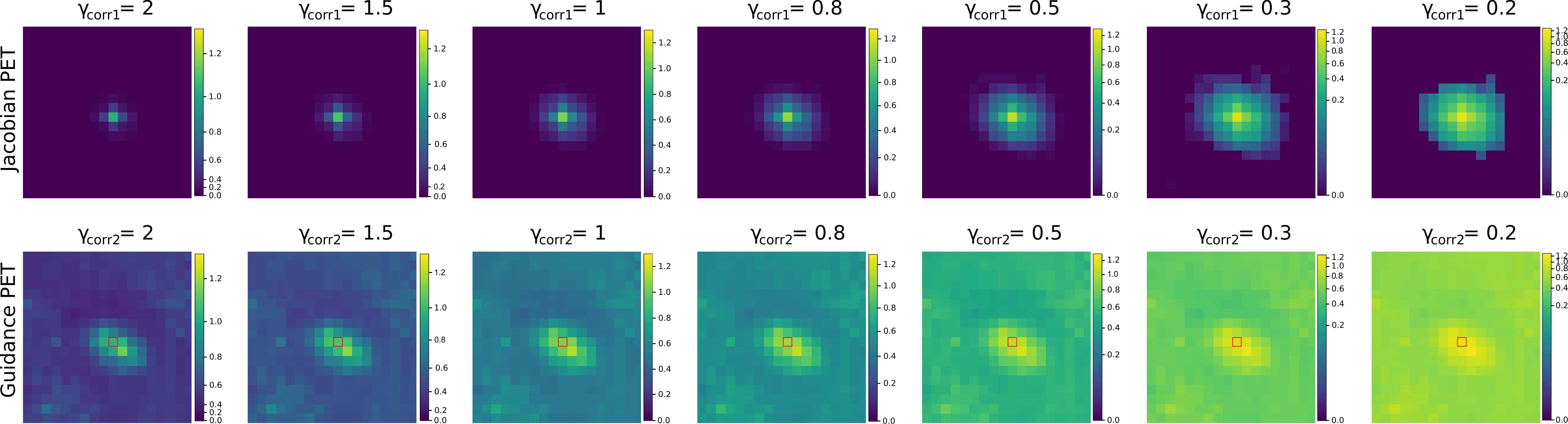}
\caption{The first and second row shows the effect of the gamma correction on the \textit{Jacobian PET} and \textit{Guidance PET} images of MaskNet network with $\gamma_{corr1}$ and $\gamma_{corr2}$ varying between 0.1 and 2.0.}
\label{fig:gamma correction}
\end{figure}

\subsubsection{Gamma correction}
The jacobian images inherently have low luminance since the gradients of the principle pixel is extremely bright compared to the gradient of its neighborhood pixels. Therefore, to better visualize the small gradient values in the jacobian images, we used the gamma correction, $\gamma_{corr1}$, on jacobian images as $\Big(\frac{\partial y^{(i)}}{\partial \boldsymbol{x}}\Big)^{\gamma_{corr1}}$. To also adjust the luminance of the guidance images, we defined another gamma correction,$\gamma_{corr2}$, on guidance images as $\Big(\frac{\partial \boldsymbol{y}}{\partial \boldsymbol{x}}\Big)^{\gamma_{corr2}}$ in a 8-bit range of RGB images. We chose two separate gamma corrections for jacobian and guidance images to independently fine-tune the luminance of these images. Gamma correction is an efficient method to improve the luminance since it only has a single parameter to fix for better visualization of the underlying gradient intensities. Currently, the range of the gamma correction values varies between $0.1$ and $2$ in our FuseVis tool. An example of the effect of $\gamma_{corr1}$ and $\gamma_{corr2}$ on jacobian and guidance images respectively is shown in Figure \ref{fig:gamma correction}. It can be seen from the images that as the $\gamma_{corr1}$ and $\gamma_{corr2}$ decreased below $1$, the luminance of the images increased due to which the small gradients became brighter while ${\gamma_{corr}}$ above 1 decreased the luminance of the small gradients.

\subsection{Overview of FuseVis tool}
An overview of the FuseVis tool is shown in Figure \ref{fig:fig1} where the images related to the MRI are placed at the top row, the images related to PET at the bottom row whereas all the modality combining images are in the middle row. We placed input and fused images on the left, guidance images in the middle, and the jacobian images on the right side of the tool. The tool also consists of a dropdown menu to select the fusion method and two separate slider widgets, namely $Gamma1$ that represents $\gamma_{corr1}$ to fix the gamma correction of the jacobian images and $Gamma2$ that represents $\gamma_{corr2}$ to fix the gamma correction of the guidance images. The principle pixel can be chosen in any of the input images, guidance images, or the fused image using the mouseover operation. The user performs the mouseover interaction by pressing and holding the left button of the mouse and hovering around the image. While performing this operation, the current mouse cursor position defines the principle pixel coordinate and the local environment around the principle pixel is displayed in the zoomed-in versions of the images which is always placed to the right of the respective images. The local environment of the principle pixel is defined based on the red squared bounding box and the current principle pixel coordinate, $i$, is positioned around the center of the bounding box which is also shown as the red square in the zoomed images. 

\begin{figure}[!htb]
\begin{subfigure}{0.25\textwidth}
  \includegraphics[width=\linewidth]{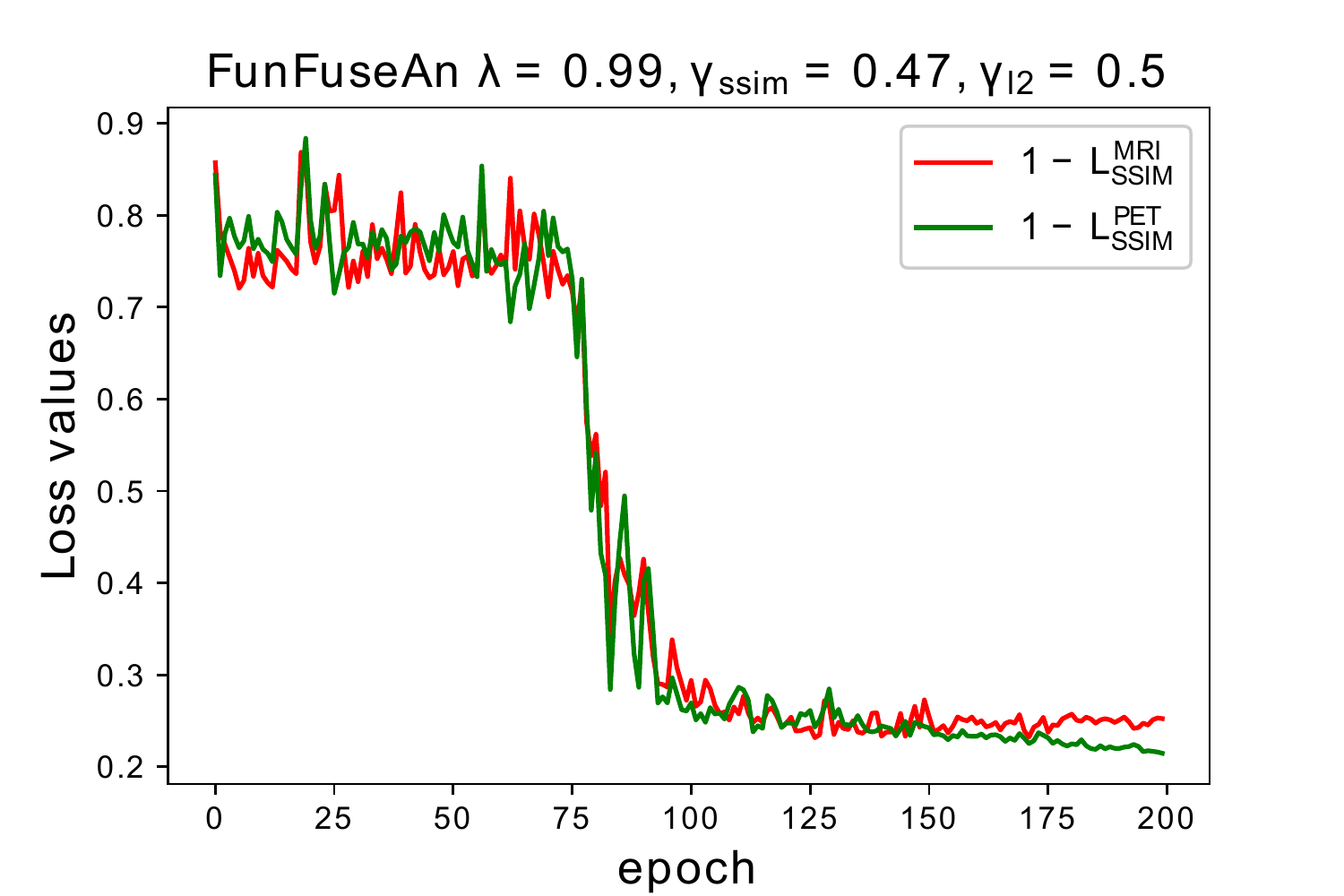}
\end{subfigure}%
\begin{subfigure}{0.25\textwidth}
  \includegraphics[width=\linewidth]{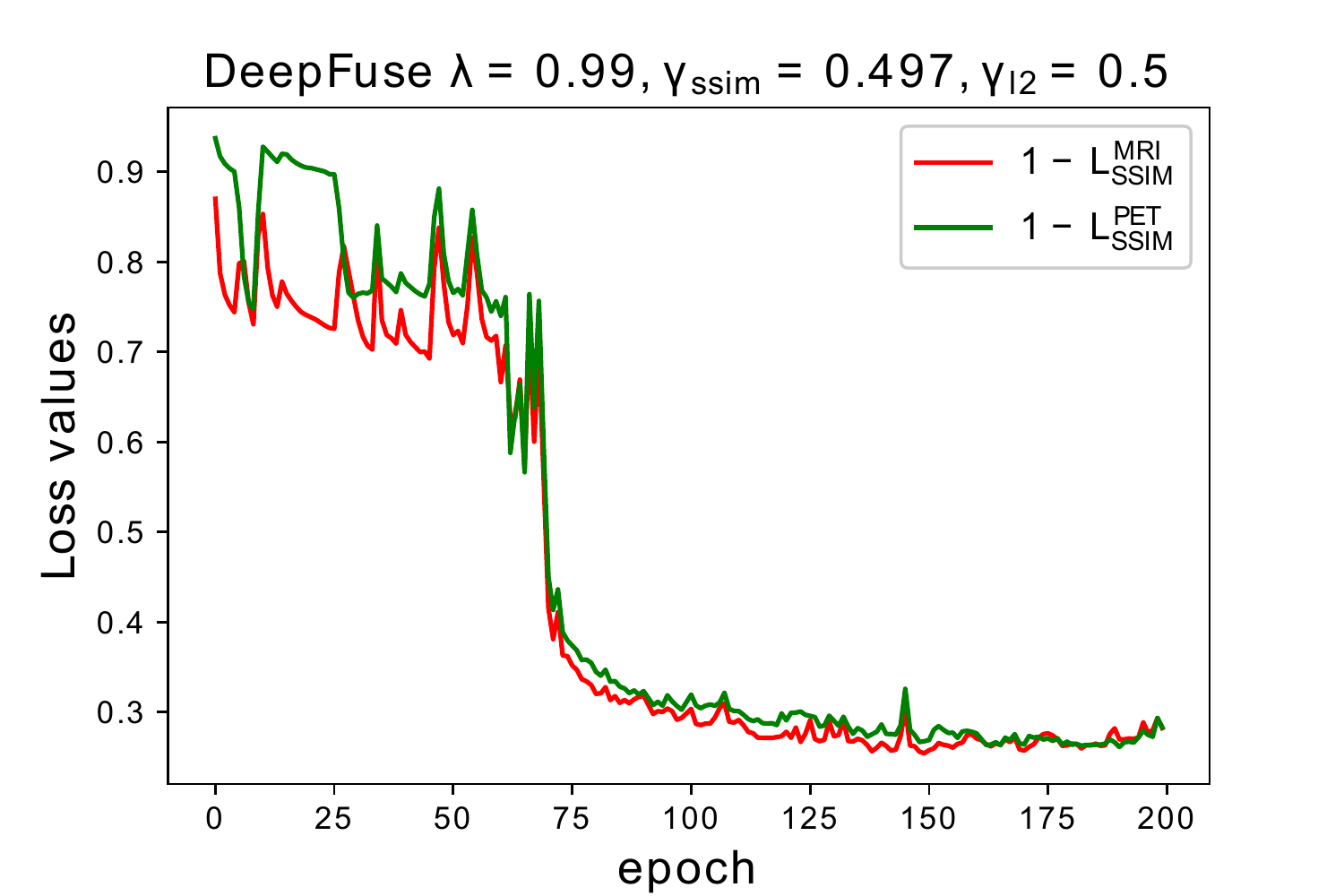}
\end{subfigure}%
\begin{subfigure}{0.25\textwidth}
  \includegraphics[width=\linewidth]{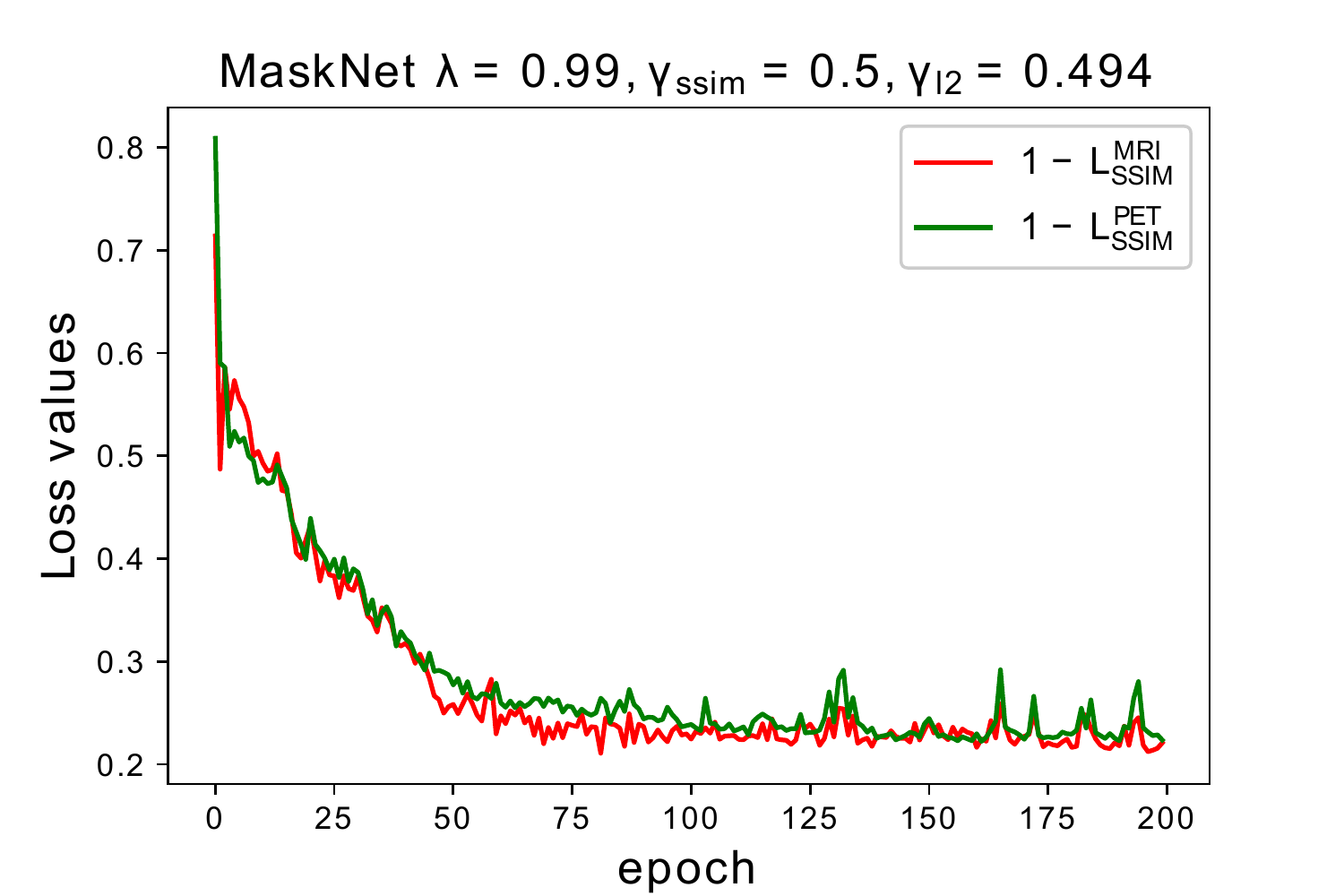}
\end{subfigure}%
\begin{subfigure}{0.25\textwidth}
  \includegraphics[width=\linewidth]{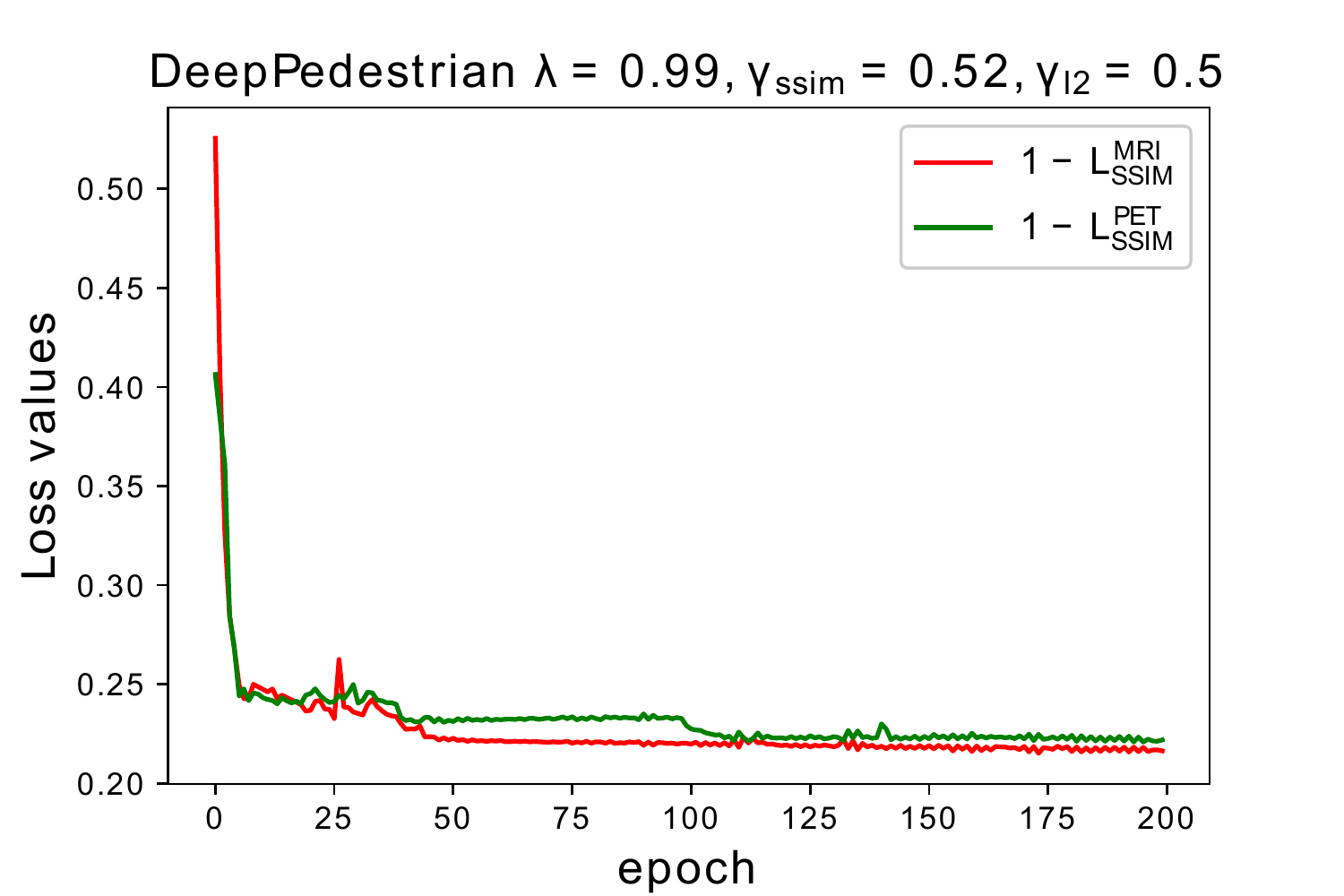}
\end{subfigure}
\\
\begin{subfigure}{0.25\textwidth}
  \includegraphics[width=\linewidth]{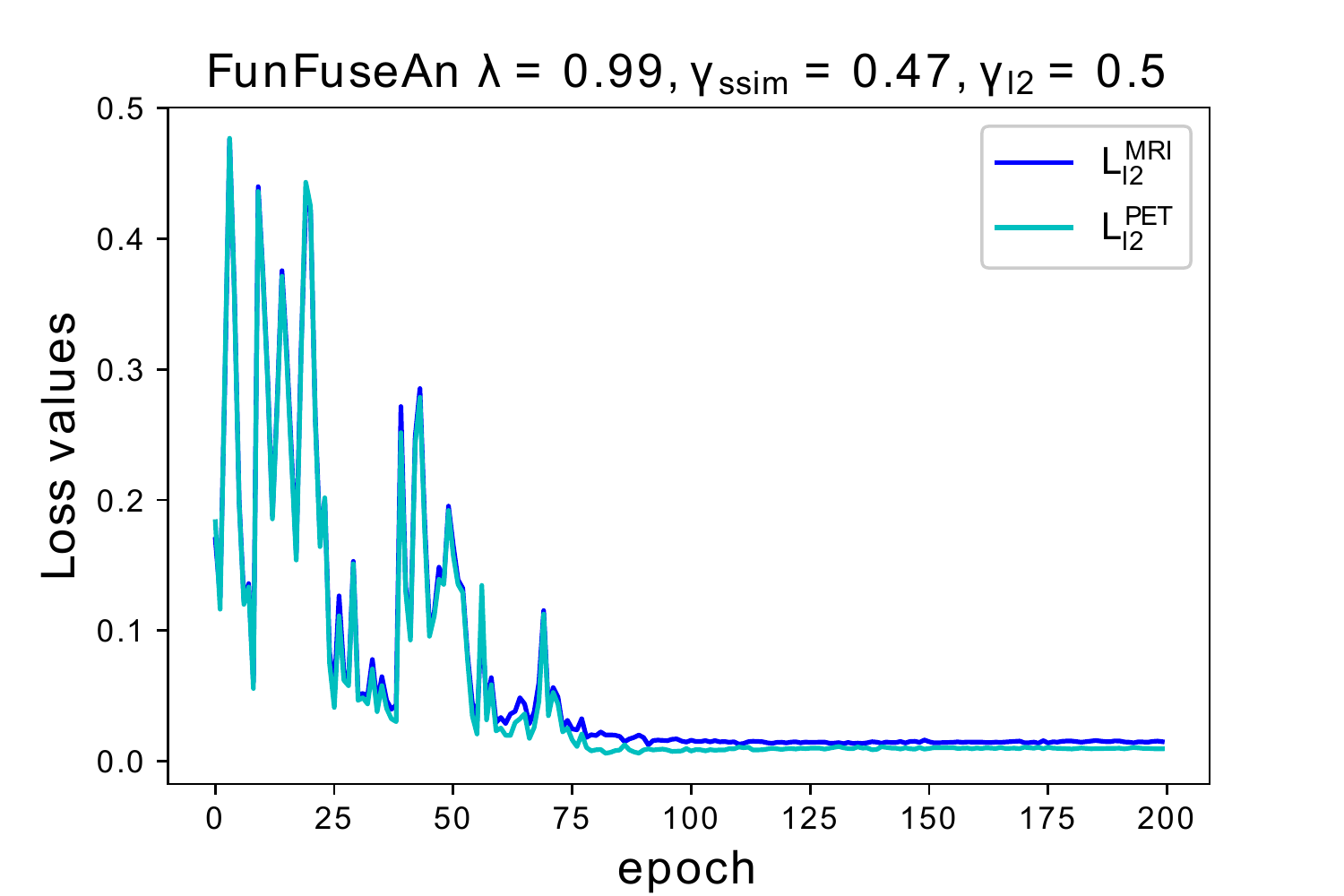}
\end{subfigure}%
\begin{subfigure}{0.25\textwidth}
  \includegraphics[width=\linewidth]{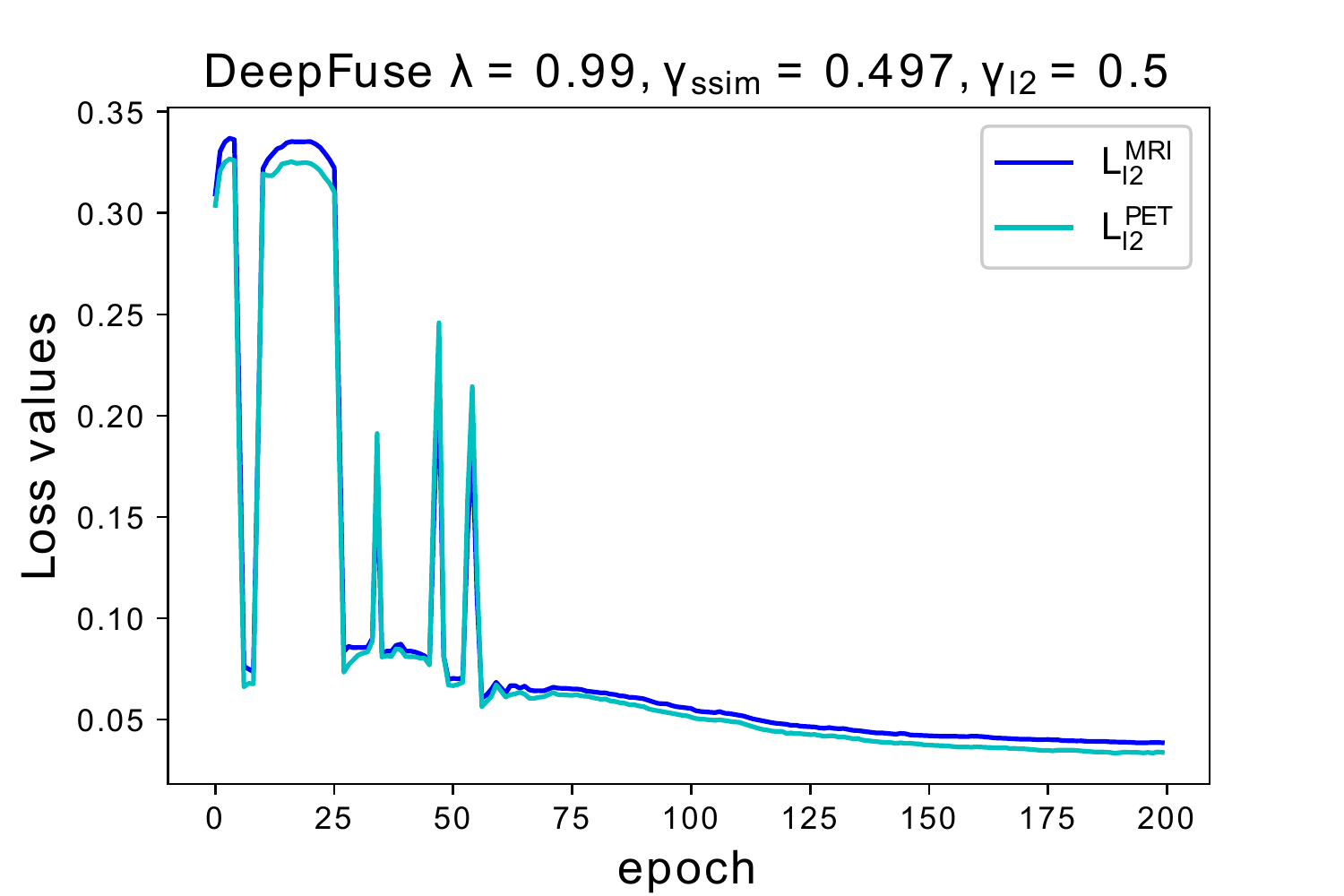}
\end{subfigure}%
\begin{subfigure}{0.25\textwidth}
  \includegraphics[width=\linewidth]{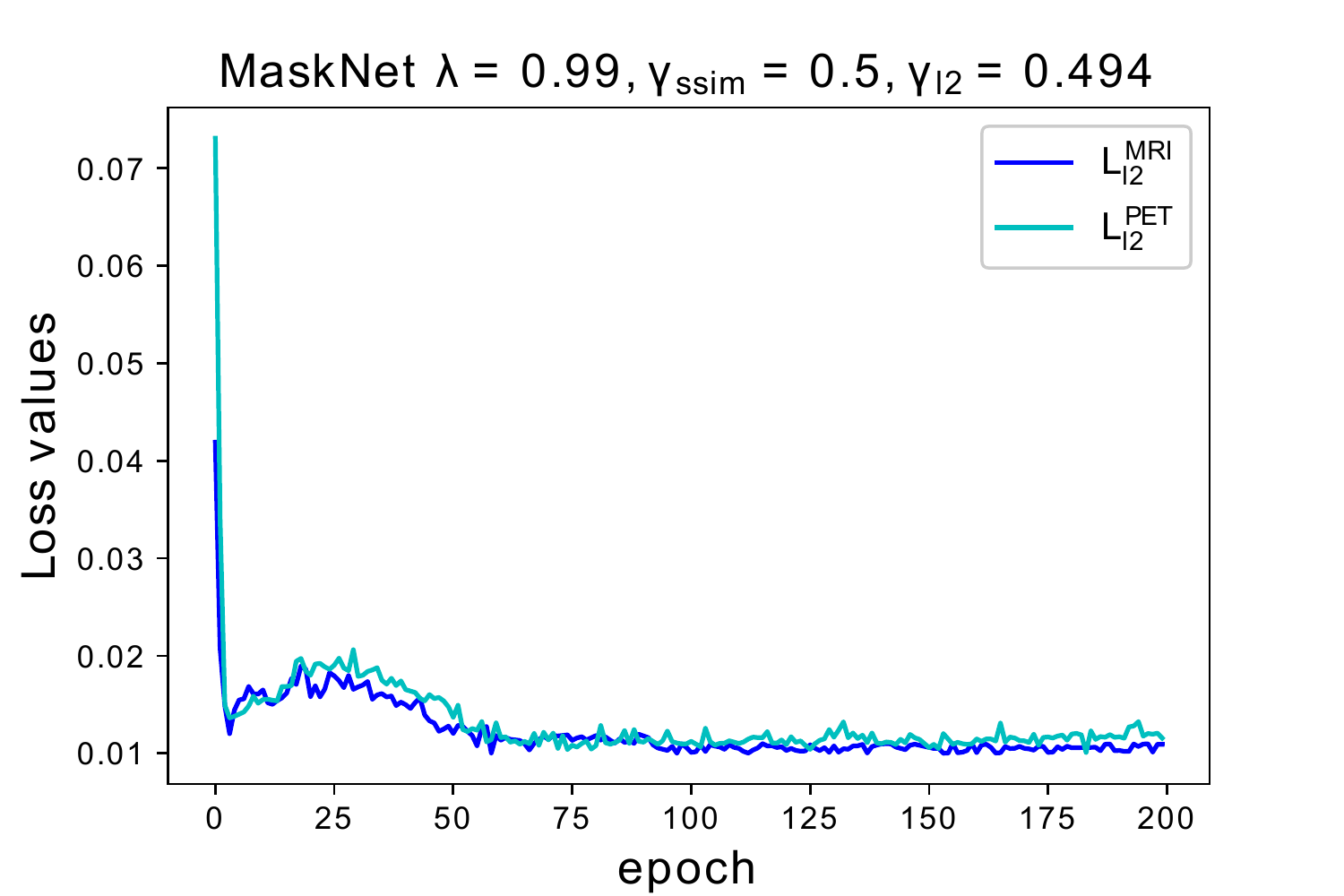}
\end{subfigure}%
\begin{subfigure}{0.25\textwidth}
  \includegraphics[width=\linewidth]{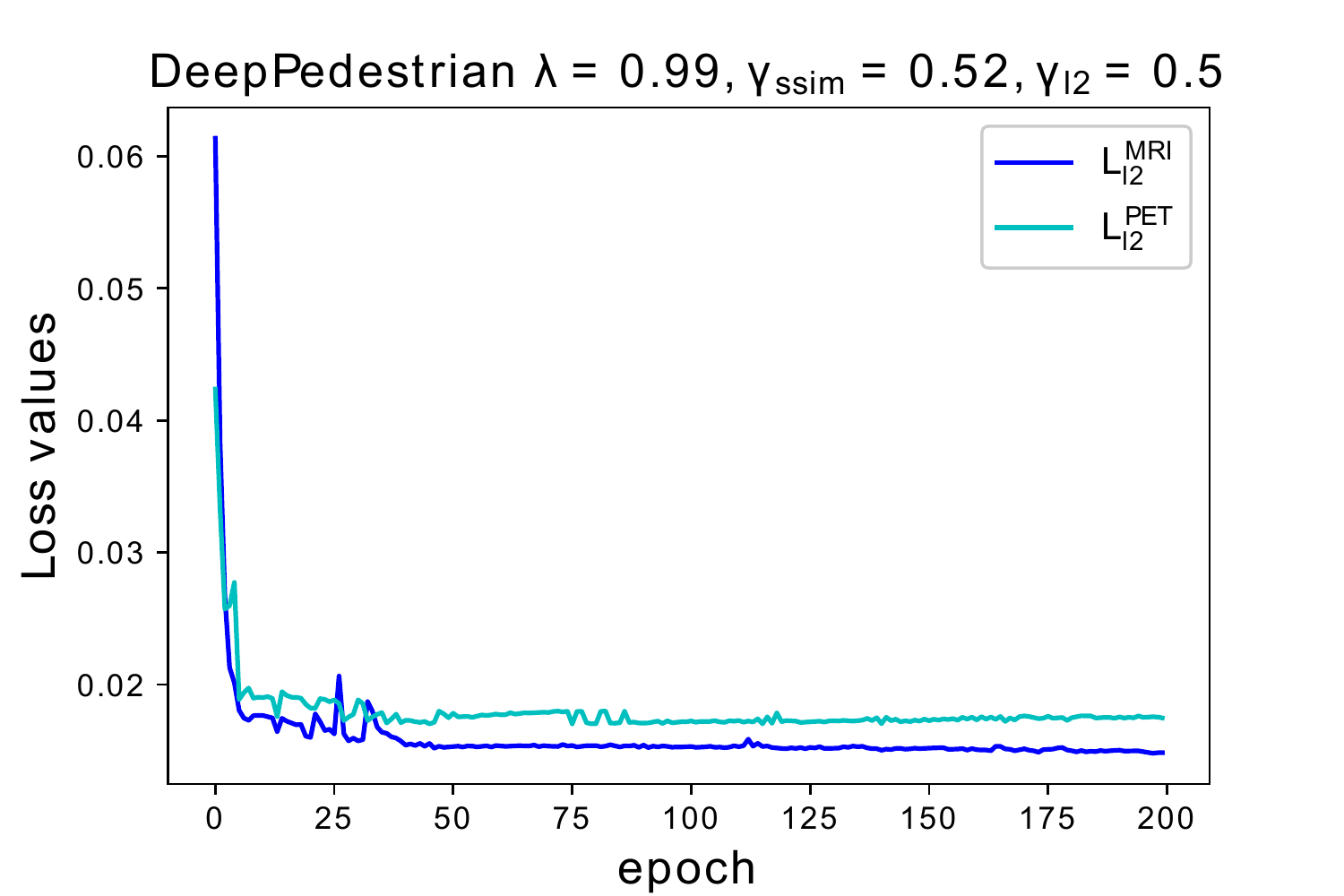}
\end{subfigure}%
\\
\begin{subfigure}{0.25\textwidth}
  \includegraphics[width=\linewidth]{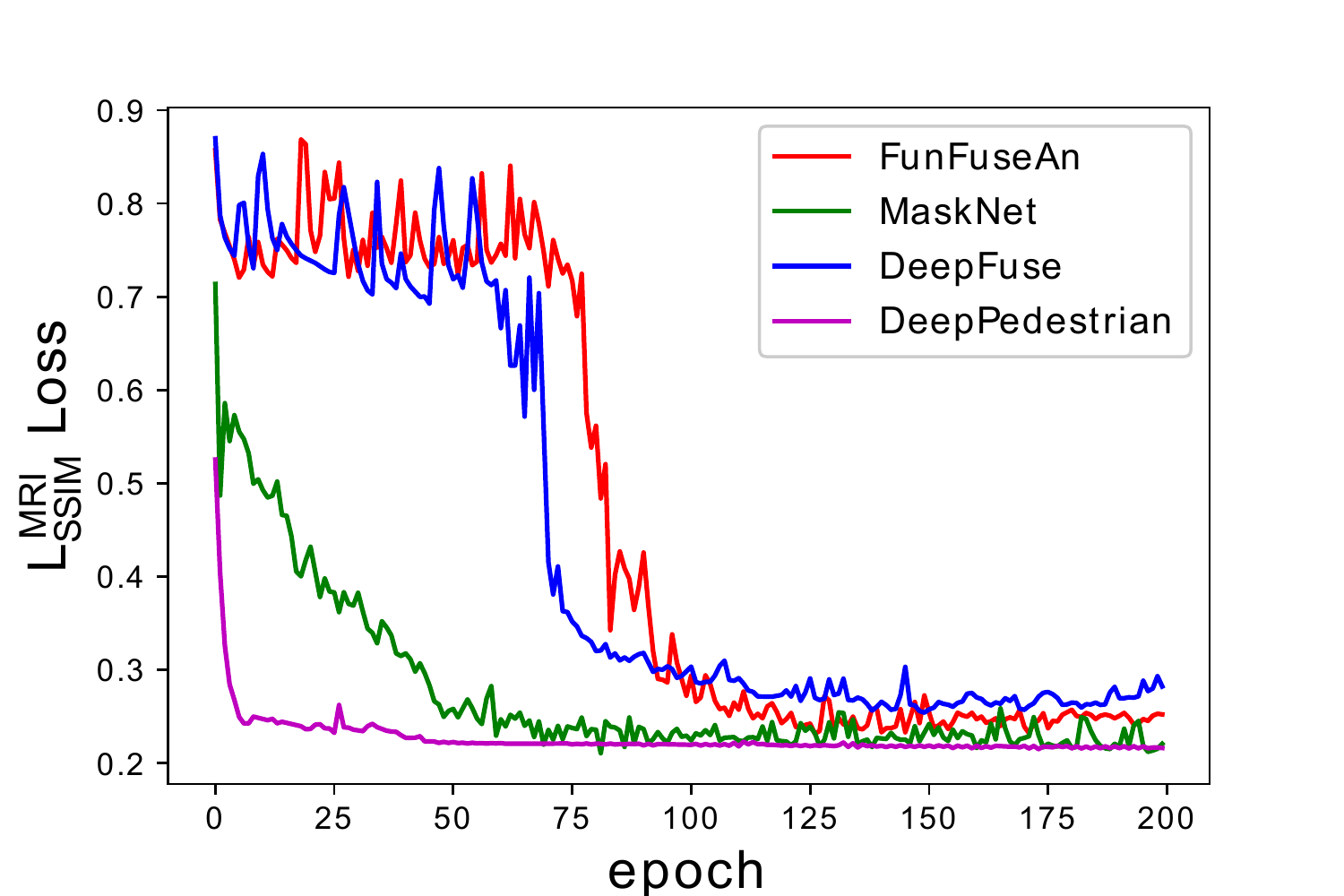}
\end{subfigure}%
\begin{subfigure}{0.25\textwidth}
  \includegraphics[width=\linewidth]{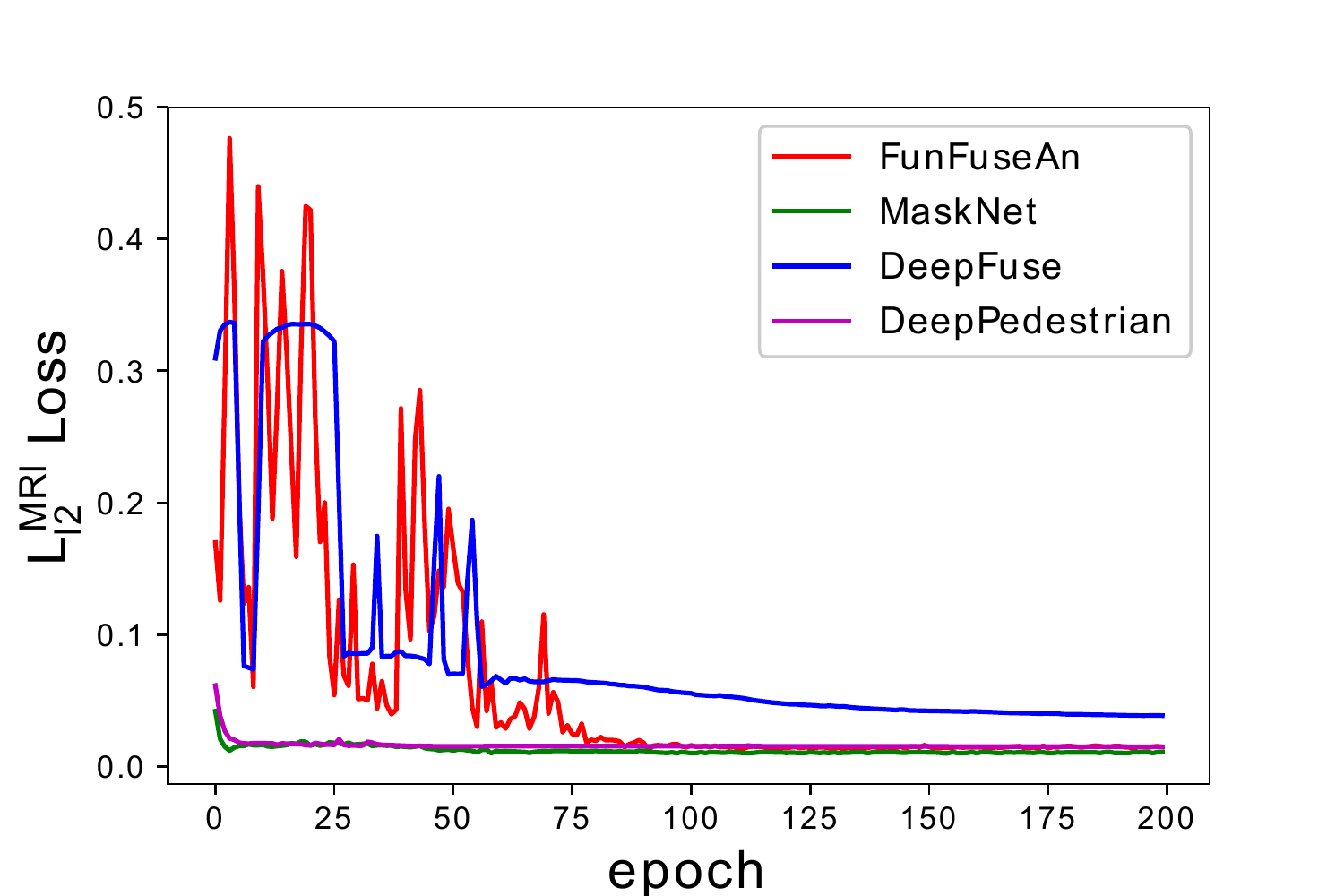}
\end{subfigure}%
\begin{subfigure}{0.25\textwidth}
  \includegraphics[width=\linewidth]{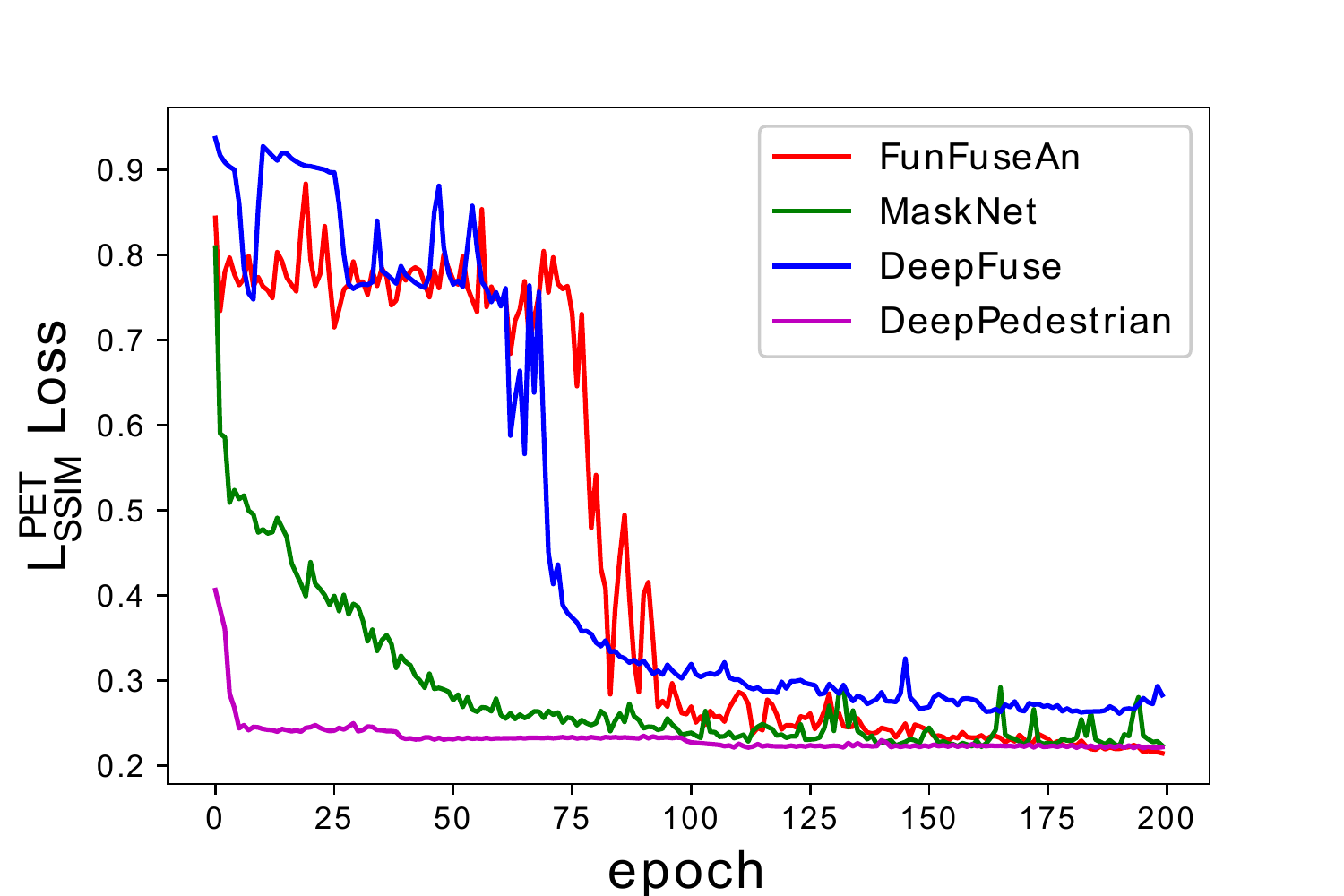}
\end{subfigure}%
\begin{subfigure}{0.25\textwidth}
  \includegraphics[width=\linewidth]{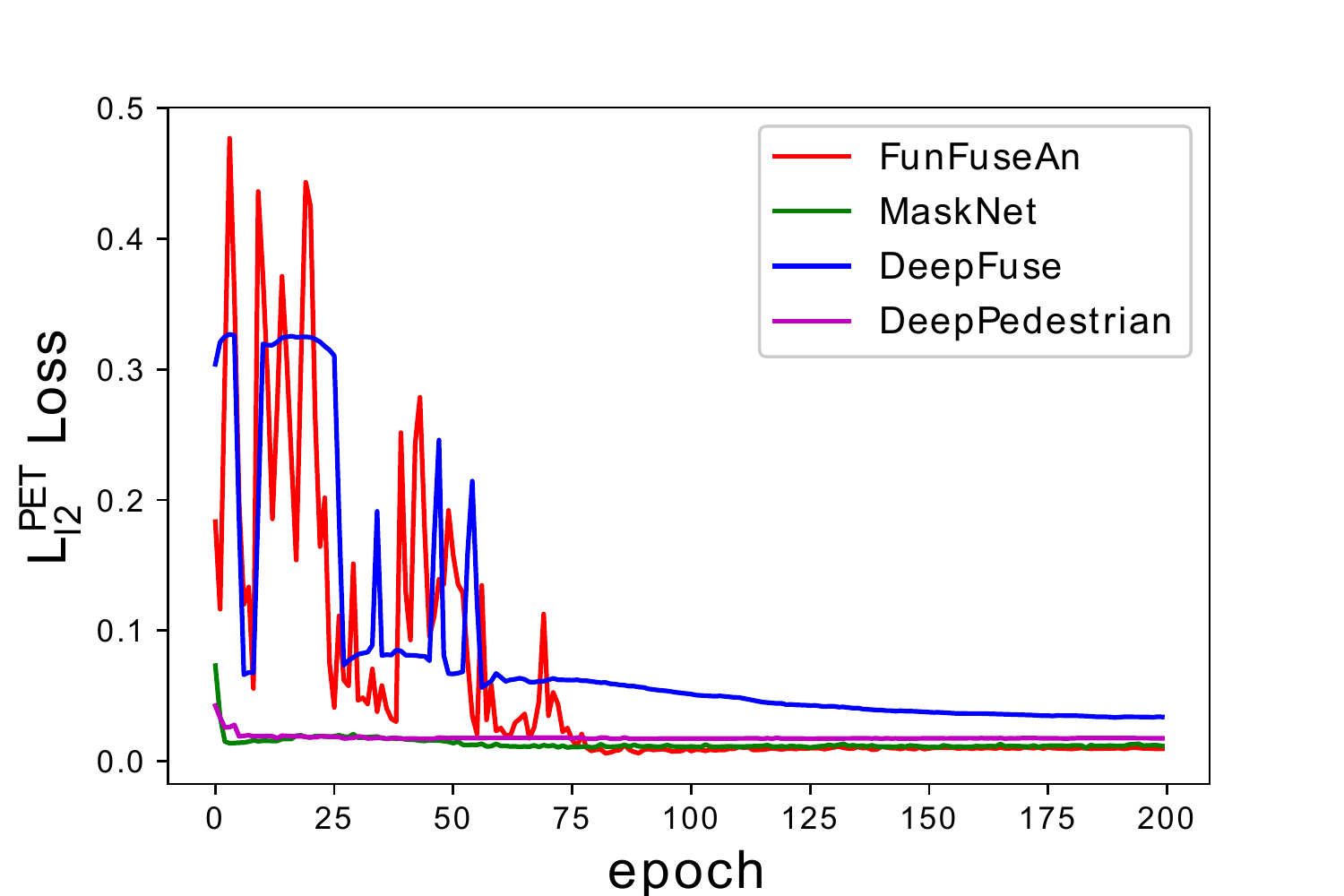}
\end{subfigure}%
\caption{Loss curves of the evaluated fusion based neural networks trained on MRI-PET image pairs.}
\label{fig:loss curves}
\end{figure} 

\section{Experimental Setup}
As there are multiple real-world applications for image fusion, it is not possible to interpret fusion based neural networks for each of these experimental settings. Therefore, we focus on the evaluation of fusion based neural networks on a specific clinical application that deals with MRI-PET fusion for tumor resection planning. For this, we first comparably train the different neural networks. Then, we present medical case studies where we mention key visualization requirements of a robust fusion method. In the same section, we select the features in the input images that are most important for the clinical diagnosis. Finally, we use our tool to find out which of the fusion methods is best suited for this specific clinical application.

\subsection{Fusion networks}
We visualized multiple end-to-end unsupervised learning based fusion approaches available in popular image fusion literature using our FuseVis tool. The image fusion based neural networks that we evaluated are \textit{FunFuseAn} \cite{ref46}, \textit{MaskNet} \cite{ref52}, \textit{DeepFuse} \cite{ref47} and \textit{DeepPedestrian} \cite{ref48}. All of these fusion based neural networks have distinct network architectures (as shown in Figure \ref{fig:networks}), built specifically for different types of image fusion applications. In addition to these four end-to-end unsupervised fusion based neural networks, we also evaluated a non-machine learning based fusion approach by implementing a simple weighted averaging technique and compared its visualization results with those from these fusion based neural networks. The weighted averaging based image fusion is a pixel-level method that assigns higher weights to the sharpest pixels in each input image. Assuming pixel $x_1^{(i)}$ from one input image to be extremely bright compared to pixel $x_2^{(i)}$ from the other input image, then the weighted averaging method will give higher weightage $w_1^{(i)}$ to $x_1^{(i)}$ and lower weightage $w_2^{(i)}$ to $x_2^{(i)}$ , thereby resulting in a bright fused pixel element $y^{(i)}$:

\begin{equation}
\label{eqn:label3}
w_1^{(i)} = \frac{x_1^{(i)}}{x_1^{(i)} + x_2^{(i)}},  \;\;\; \; w_2^{(i)} = \frac{x_2^{(i)}}{x_1^{(i)} + x_2^{(i)}}, \;\; \; \; y^{(i)} = w_1^{(i)}*x_1^{(i)} + w_2^{(i)}*x_2^{(i)}
\end{equation}

\subsection{Hardware-software setup}
 The work was performed with Python (version 2.7) and deep learning library PyTorch was used to implement fusion based neural networks. The Python wrapper of the GUI library \textit{Tkinter} was used to create the user interface of our FuseVis tool. All the visualization results were derived from FuseVis tool using a 64 bit Windows operating system equipped with a single GeForce GTX 1080 Ti GPU and 12 Intel Core i7-8700K CPU @ 3.70 GHz with 64-GB RAM. The same hardware-software setup was used for the real-time computation of the jacobian images. However, to fasten the training of the neural networks and to speed up the static computations of guidance images, we used a more powerful hardware setup containing six NVIDIA Tesla V100 GPUs. 

\subsection{Training dataset}
For training each of the fusion networks in a uniform setup, we obtained several MRI-T2 and PET image pairs of 50 different patients publicly available at the Alzheimer’s Disease Neuroimaging Initiative (ADNI) \cite{ref73} with the age of patients varying between 55-90 years among both genders. The patients were chosen in such a way that the training dataset covers mild, moderate, and severe stages of Alzheimer's disease. All images were extracted as axial slices with a voxel size of 1.0 x 1.0 x 1.0 mm$^3$. The MRI-T2 images were N3m MPRAGE sequences while PET images were co-registered, averaged, and standardized with a uniform resolution for each of the subjects. We aligned the MRI-PET image pairs using the Affine transformation tool of the 3D Slicer registration library. 

\subsection{Network architectures}
The architectures of the evaluated fusion based neural networks are shown in Figure \ref{fig:networks}. A \textit{Conv 3 x 3 x 9 x 16} block, for example, means a convolution operation performed on an input with nine channels that results in an output with 16 channels. For this, 16 kernel filters each with size \textit{3 x 3} and nine channels are utilized. Each channel of the kernel filter is convoluted with the corresponding input channel and subsequently summed up to generate a single feature map. This operation is repeated for all the kernel filters to extract output feature maps equal to the number of filters. The weights of the kernel filter have been initialized based on a uniform distribution while the striding and the padding operation have been performed in such a way that there is no downsampling in any hidden layer. A \textit{BatchNorm} means a batch normalization operation and \textit{LReLU} means a leaky ReLU activation function with a slope 0.2. It is noticeable that each of the networks has different architectures due to varying image fusion tasks that they solve. The architecture of \textit{FunFuseAn} \cite{ref46} is inspired by the classical non-machine learning based image fusion technique of feature extraction, fusion, and feature reconstruction with specific application to the MRI and PET image pairs. \textit{MaskNet} \cite{ref52} concatenates the input images and performs a dense feature fusion in the initial layers of the network before constructing a fused image. \textit{DeepFuse} \cite{ref47} defines an architecture where the feature extraction layers have tied weights for each of the input image modalities. This helps in increasing the brightness of an underexposed input image by using the luminance of an overexposed input image. \textit{DeepPedestrian} \cite{ref48} also concatenated the input images and utilized a deep architecture with multiple residual blocks specifically for infrared and visible image fusion. Since we aim to interpret the behavior of these image fusion based neural networks by performing a per-pixel gradient based saliency visualization using our FuseVis tool, we modified the original training setup of these fusion networks and defined a common strategy to train these networks.

\subsection{Loss Function}
An image fusion approach aims to preserve as much of the input image features as possible in the fused image. Therefore, the training objective should be defined in such a way it could measure the similarity between the input and the fused image and optimize the network parameters in order to achieve perceptually similar results. 
Structural Similarity Index (SSIM) \cite{ref9} is an efficient similarity metric that is also differentiable due to which it can be adopted as a primary loss function for training each of the fusion based neural networks. However, empirically setting only SSIM as the loss function leads to the change of brightness in the final fused image. The $\ell_2$ loss helps to overcome this shortcoming by preserving the luminance component better. Therefore, we define a combination of SSIM and $\ell_2$ losses to train each of the neural networks as shown in Equation \ref{eqn:ref12}. The $\boldsymbol{x_1}$ and $\boldsymbol{x_2}$ are MRI and PET training images respectively, the regularization hyperparameter $\lambda$ controls the weightage of the total $L_{SSIM}$ and total $L_{\ell_2}$ losses while the hyperparameters $\gamma_{ssim}$ and $\gamma_{\ell_2}$ controls the weightage between the individual SSIM and $\ell_2$ losses of each input image modality respectively. 

\begin{equation}
\label{eqn:ref12}
\begin{gathered}
L_{SSIM}^{MRI} = 1 - SSIM(\boldsymbol{x_1}, \boldsymbol{y}),  \;\;\;\;  L_{SSIM}^{PET} = 1 - SSIM(\boldsymbol{x_2}, \boldsymbol{y})\\
L_{SSIM} = \gamma_{ssim}*L_{SSIM}^{MRI} + (1-\gamma_{ssim})*L_{SSIM}^{PET} \\
L_{\ell_2}^{MRI} = ||\boldsymbol{y} - \boldsymbol{x_1}||_2, \;\;\;\; L_{\ell_2}^{PET} = ||\boldsymbol{y} - \boldsymbol{x_2}||_2 \\
L_{\ell_{2}} = \gamma_{\ell_2}*L_{\ell_2}^{MRI} + (1-\gamma_{\ell_2})*L_{\ell_2}^{PET} \\
 L_{total} = \lambda * L_{SSIM} + (1 - \lambda) * L_{\ell_2}
\end{gathered}
\end{equation}

The loss convergence curves of all the evaluated unsupervised learning based fusion networks are shown in Figure \ref{fig:loss curves}. Each of the fusion networks was trained for 200 epochs, the batch size was fixed as 2 and the Adam optimizer was used as the optimization function during the backpropagation step with a learning rate of $2 x 10^{-3}$. 

\subsection{Hyperparameter tuning}
Ideally, a fused image should preserve features from both the input images equally and there should be minimal addition of the brightness artifacts. For achieving this, we followed a hyperparameter tuning strategy where we aimed to find a plausible balance between the $L_{SSIM}^{MRI}$ and $L_{SSIM}^{PET}$ as well as $L_{\ell_2}^{MRI}$ and  $L_{\ell_2}^{PET}$ loss curves by training a single fusion network with multiple combinations of $\lambda$, $\gamma_{ssim}$ and $\gamma_{\ell_2}$ values. Once a good balance was found between these curves, we evaluated the best $\lambda$ configuration that leads to the least brightness artifacts by analyzing the resultant fused images.
We initially trained the \textit{FunFuseAn} network with $\lambda$ = 0.01, 0.2, 0.5, 0.8 and 0.99 with $\gamma_{ssim}$ and $\gamma_{\ell_2}$  each equal to 0.1, 0.3, 0.5, 0.7 and 0.9. Therefore, a total of 125 instances of \textit{FunFuseAn} network were learned in separate training environments. For each $\lambda$ configuration, we analyzed the combination of $\gamma_{ssim}$ and $\gamma_{\ell_2}$ that nearly balances the SSIM and $\ell_2$ loss curves for both MRI and PET images. Then, we fine tuned $\gamma_{ssim}$ and $\gamma_{\ell_2}$ parameters for each $\lambda$ by training several more instances of the \textit{FunFuseAn} network in the vicinity of the previous best $\gamma_{ssim}$ and $\gamma_{\ell_2}$ in order to find the next best configuration that balances the loss curves even better.

\begin{table}[H]
\caption{The table shows the partial loss values of the trained fusion networks after 200 epochs and the fine tuned hyperparameter configurations.} 
\centering
\begin{tabular}{|P{2.5cm}|P{1cm}|P{1cm}|P{1cm}|P{1.3cm}|P{1.3cm}|P{1.3cm}|P{1.3cm}|}
\hline Network & $\lambda$ & $\gamma_{ssim}$ & $\gamma_{\ell_2}$ &  $L_{SSIM}^{MRI}$ & $L_{SSIM}^{PET}$ & $L_{\ell_2}^{MRI}$ & $L_{\ell_2}^{PET}$ \\
\hline
FunFuseAn & 0.99 & 0.47 & 0.5 & 0.2524  & 0.2147 & 0.0148 &  0.0094 \\
\hline
MaskNet	& 0.99 & 0.5 & 0.494 & 0.2208 & 0.2236 & 0.0109 & 0.0115 \\
\hline
DeepFuse & 0.99 & 0.497 & 0.5 & 0.2824  & 0.2830 & 0.0385 & 0.0338\\
\hline
DeepPedestrian & 0.99 & 0.52 & 0.5 & 0.2164  & 0.2219 & 0.0148 & 0.0175 \\
\hline
\end{tabular}
\label{tab:hyper}
\end{table}

 After fixing the best $\gamma_{ssim}$ and $\gamma_{\ell_2}$ configuration for each $\lambda$, we determined which of the $\lambda$ values leads to negligible brightness artifacts in the fused image. Therefore, we trained new network instances with additional values of $\lambda$ =  0.9, 0.999, 0.9999, 0.99999, 0.999999 and 1.0  to properly evaluate the corner cases of our loss function. We observed that the brightness artifact in the fused images worsened as the values of $\lambda$ were increased above 0.999 with $\lambda$ = 1.0 having the most intense brightness in the fused image. $\lambda$ = 0.99 was one configuration where we observed negligible change in brightness of the fused image which meant a weightage of 0.01 to $\ell_2$ was sufficient to remove the brightness artifact problem from the SSIM loss function. $\lambda$ = 0.999 also had negligible brightness addition but the fused image from this configuration lost some PET features due to which we fixed $\lambda$ = 0.99 as the best hyperparameter for the $FunFuseAn$ network. Since $\lambda$ = 0.99 was sufficient to overcome the brightness artifact problem, we used the same $\lambda$ = 0.99 value for all the other fusion networks and then evaluated several instances of these networks to obtain the best $\gamma_{ssim}$ and $\gamma_{\ell_2}$ that balances the loss curves. The loss curves in Figure \ref{fig:loss curves} were generated with the best hyperparameter configuration for each of the fusion based neural networks. The final hyperparameter values for the fusion networks can be seen in Table \ref{tab:hyper}. It is observable that all the networks were able to balance the loss curves while \textit{DeepFuse} network converged at higher loss values compared to the other fusion networks.


\section{Medical Case studies}
In this section, we discuss the specific clinical application of the MRI-PET image fusion in detail. Then, we perform case studies on pre-registered MRI and PET image pairs by providing the pathological information about the patients from whom these images were acquired. Finally, we will present key visualization requirements that will make a fusion approach suitable for usage in this specific clinical application.    

\subsection{Glioma and its pathological features}
Glioma is a type of brain tumor that originates in the glial cells that surround and support the neurons in the brain. MRI enables the clinicians to estimate the size and the location of glioma, thereby acting as an initial marker for malignancy. For example, in an MRI-T2 grayscale image, brighter pixel intensities imply hyperintense signal abnormalities which are observable in the brain regions with a tumorous lesion such as glioma. The grayscale intensities in these tumorous lesions could be similar to the grayscale intensities in the lateral ventricle region of the brain that mostly contains cerebrospinal fluid (CSF) with high water content. However, there could be several sub-regions within the boundary of high-grade glioma with varying pathological features, which is challenging to visualize by only interpreting the MRI images as the MRI contrast is quite uniform within the tumor boundary. Hence, the aggressiveness and increased potential of the sub-tumor tissues cannot be adequately determined.

Functional imaging modalities such as PET and SPECT are better equipped to differentiate between the sub-regions of glioma by visualizing the functional information like glucose metabolism and the extent of CBF or perfusion activity in these sub-regions. The innermost part of glioma is the necrotic core that contains dead tissues and there is no possibility of CBF and/or glucose metabolism due to which it has very dark PET features. The region that surrounds necrotic tissues is called an enhancing tumor which is filled with inflammation fluid and the blood-brain barrier leaks in this region due to cancerous angiogenesis. Hence, this region has glucose metabolism with blood supply to support the cancerous cell growth resulting in lighter dark PET features. The necrotic core and enhancing tumor region constitute the bulk of glioma that clinicians want to visualize using fused MRI and PET features and performs a precise resection of these two regions. The non-enhancing solid tumor is the outermost region of glioma which is generally not resected since the blood-brain barrier is still intact in these tissues. The bright regions in the PET image resemble healthy tissues with a high level of blood perfusion and normal glucose metabolism.

\subsection{Clinical test examples}
The clinical test examples were acquired from Harvard Whole Brain Atlas database \cite{ref74} with the combination of MRI-T2 and PET/SPECT images from patients suffering from different types of glioma. The test examples were disjoint to the training dataset and the network has not seen such clinical pathology during its training. The first clinical MRI-PET image pair is shown in the first four columns of Figure \ref{fig:fused}. The scans are of a patient who was suffering from \textit{Anaplastic Astrocytoma}, a rare and malignant brain lesion classified under the category of high-grade glioma.  A lesion in the right and the left side of the brain is visible in the MRI image and has bright grayscale intensities. On the other hand, bright regions in the PET image convey normal blood flow while very dark regions suggest no blood flow to the necrotic tissues. In the second clinical MRI-PET image pair as shown in the last four columns in Figure \ref{fig:fused}, the patient had a long history of tobacco usage and was originally suffering from \textit{Metastatic Bronchogenic Carcinoma}, which is a type of lung cancer. The patient began having headaches and the scans revealed brain metastases that occurred due to the spread of cancer cells present in the lungs to the brain resulting in the diagnosis of glioma. A lesion in the right side of the brain with bright features is visible in the MRI image while very dark PET features reveal no blood flow in the necrotic region.

\subsection{Visualization requirements}
\begin{figure}[!htb]
\centering
\includegraphics[width=15cm]{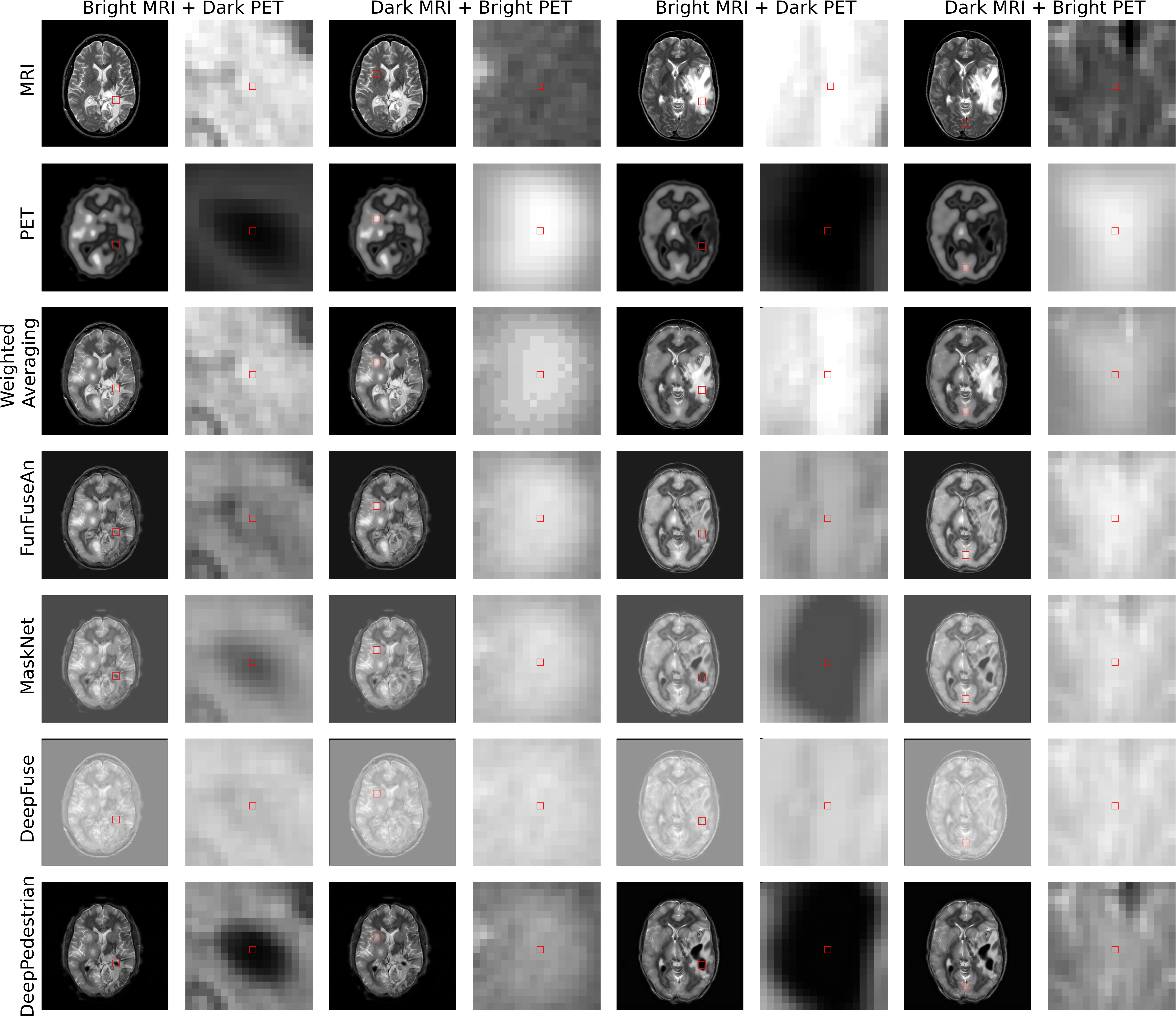}
\caption{The figure shows the fusion results for each of the fusion methods. The zoomed image within the clinical region of interests are always placed on the right of the unzoomed image.}
\label{fig:fused}
\end{figure}

As the anatomical MRI image cannot precisely estimate the sub-tumor boundaries within glioma and the functional PET/SPECT image is unable to model the overall extent of the tumor; therefore, one of the goals of MRI-PET image fusion is to better delineate the sub-peripheries of glioma by envisioning the metabolic and/or CBF characteristics of its sub-regions. However, it is challenging to interpret the suitability of these fused images for the mentioned clinical application, if the results are not supported by supplementary visualization tools. This secondary visual analysis becomes critically important when several fusion methods provide similar results. Additionally, the recent image fusion methods are developed using unsupervised learning based deep neural networks that are non-transparent (i.e. a ‘black box’) due to its difficulty to retrace the fusion decision in light of huge parameter space. This shortcoming is especially problematic in sensitive medical domains like MRI-PET image fusion where understanding and explaining fusion results obtained from neural networks is required for a robust clinical decision. Therefore, the following visual tools are required to understand the suitability of a fusion approach for usage in the clinical application:
\begin{itemize}
    \item A fusion approach should assist clinicians in visualizing the extent of hyper dark PET regions resembling necrotic core with no blood flow being superimposed on the bright anatomical boundary of the whole tumor mass. This information is important for clinicians to estimate the extent up to which a tumor resection is required. For example, in the first and the fifth column of Figure \ref{fig:fused}, the principle pixel in this very dark PET region was chosen for visual analysis.

\item A fusion approach should preserve the very bright PET features which convey
high blood perfusion and normal metabolism in healthy brain tissues as it helps clinicians in visualizing the regions with high brain activity due to external stimuli at a particular time. For example, in the third and seventh column of Figure \ref{fig:fused}, the principal pixel in the bright PET region was chosen for visual analysis.

\item A fusion approach should be stable and less sensitive to changes in input features from a clinically less significant modality. For example, the change in grayscale MRI intensities within the necrotic core shall not highly influence the fused grayscale intensities as it might corrupt the clinically important dark PET features. For example, the \textit{MaskNet} and \textit{DeepPedestrian} networks are less sensitive to changes in the MRI features which can be visualized in the guidance MRI and guidance PET images of these networks shown in Figure \ref{fig:guidance}.

\item A fusion approach should be less sensitive to the changes in grayscale pixel intensities located in one sub-region of
glioma (say enhancing tumor) when the principle pixel is located in the other regions of glioma (say necrotic core). Therefore, a fusion method should have a negligible influence of the
neighborhood pixels exterior to a local feature with the principle pixel interior to the local feature. For example, the fusion methods such as \textit{Weighted Averaging} and \textit{MaskNet} have no or very low gradients in the neighborhood pixels which are outside the very dark PET features resembling necrotic core as shown in the jacobian images in Figure \ref{fig:jacobian}.

\end{itemize}

\section{Results and discussion}
In this section, we present a detailed visual analysis of the fusion based neural networks using our FuseVis tool. First, we will analyze the fused images from each of the fusion methods and discuss their clinical significance. Then, we will discuss the visualization insights by performing the saliency analysis and examine the feasibility of the fusion methods for usage in the clinical application. Finally, we will present the timing results of our FuseVis tool and show the tool's real time capabilities.  

\subsection{Fused images}
The fused images of the \textit{Weighted Averaging} method as shown in the third row of Figure \ref{fig:fused} gives very high weightage to the bright input intensities in the regions where intensities in the other modality are darker. Therefore, we can perceive a reproduction of features from relatively brighter regions of the MRI and PET images due to which \textit{Weighted Averaging} method is unable to preserve the clinically important dark PET features related to the necrotic core but favorably preserves the bright PET features resembling healthy tissues. The fused images of the \textit{FunFuseAn} as shown in the fourth row of Figure \ref{fig:fused} are comparatively similar to the \textit{Weighted Averaging} approach where it is not able to preserve the very dark PET features. However, the fused images from \textit{FunFuseAn} look relatively dull compared to the fused images from \textit{Weighted Averaging} since the \textit{FunFuseAn} network mixes both MRI and PET features even though the boundary information about the necrotic region is lost.

The analysis of the fused images from \textit{MaskNet} in the fifth row of Figure \ref{fig:fused} shows a significant loss of anatomical edges from MRI such as the brain skull. However, contrary to \textit{Weighted Averaging} and \textit{FunFuseAn}, \textit{MaskNet} preserved the PET features better in both the dark and bright regions resembling the necrotic core and healthy tissues respectively. However, the dark PET features are slightly blurred due to changes in the overall brightness of the fused image. The fused image obtained from \textit{DeepFuse} as shown in the sixth row of Figure \ref{fig:fused} has an even higher shift in grayscale intensities due to which the method is unable to preserve the dark PET features even though the overall anatomical MRI structures are well preserved. The change in brightness can be explained by the fact that the $L_{\ell_2}^{MRI}$ and $L_{\ell_2}^{PET}$ converged at higher loss values compared to other networks due to which the brightness component of SSIM was not properly optimized. Additionally, the architecture of \textit{DeepFuse} network has been crafted for adding exposure to underexposed images by using the brightness component from each of the input image modalities and employ it to generate very bright fusion results. The fused image from \textit{DeepPedestrian} as shown in the seventh row of Figure \ref{fig:fused} clearly delineates the boundary of the necrotic core by preserving the very dark PET features, which is of high clinical significance for medical professionals. However, the anatomical edges from MRI are lost in the fused image which is also one of the main artifacts in the \textit{MaskNet} network. Another important observation is that the bright PET features resembling healthy tissues appear to be not well preserved due to an overall brightness shift.

\begin{figure}[!htb]
\centering
\includegraphics[width=15cm]{Guidance_images.pdf}
\caption{The figure shows the guidance MRI and guidance PET images for each of the fusion methods in the clinical region of interests. The $\gamma_{corr2}$ was fixed at 0.5 for all the guidance images.}
\label{fig:guidance}
\end{figure}

\textit{Summary}: The analysis of the fused images shows that even though SSIM was used as a natural loss function for training the fusion networks in a common training setup, each of the fusion methods displayed very different fusion results. The \textit{MaskNet} and \textit{DeepPedestrian} methods provide clinically relevant results where the boundaries of the necrotic core of glioma are perceivable with the dark PET features clearly superimposed on the anatomical tumor features of MRI in the fused image. However, these methods do not provide loss-free fusion results as there are missing anatomical information in both these approaches while the dark PET features are not entirely preserved in the \textit{MaskNet} network. Crucially, it is not clear by only looking into the fused images of these two methods, that up to what extent the changes in the bright MRI features can affect the final fusion results within the clinically significant dark PET regions. This analysis is important for visualizing the stability of these networks to changes in the MRI features which is a critical visualization requirement. Hence, we will perform this analysis using the visualization concepts in the subsequent sub-sections.


\subsection{Guidance images}
\begin{figure}[!htb]
\centering
\includegraphics[width=15cm]{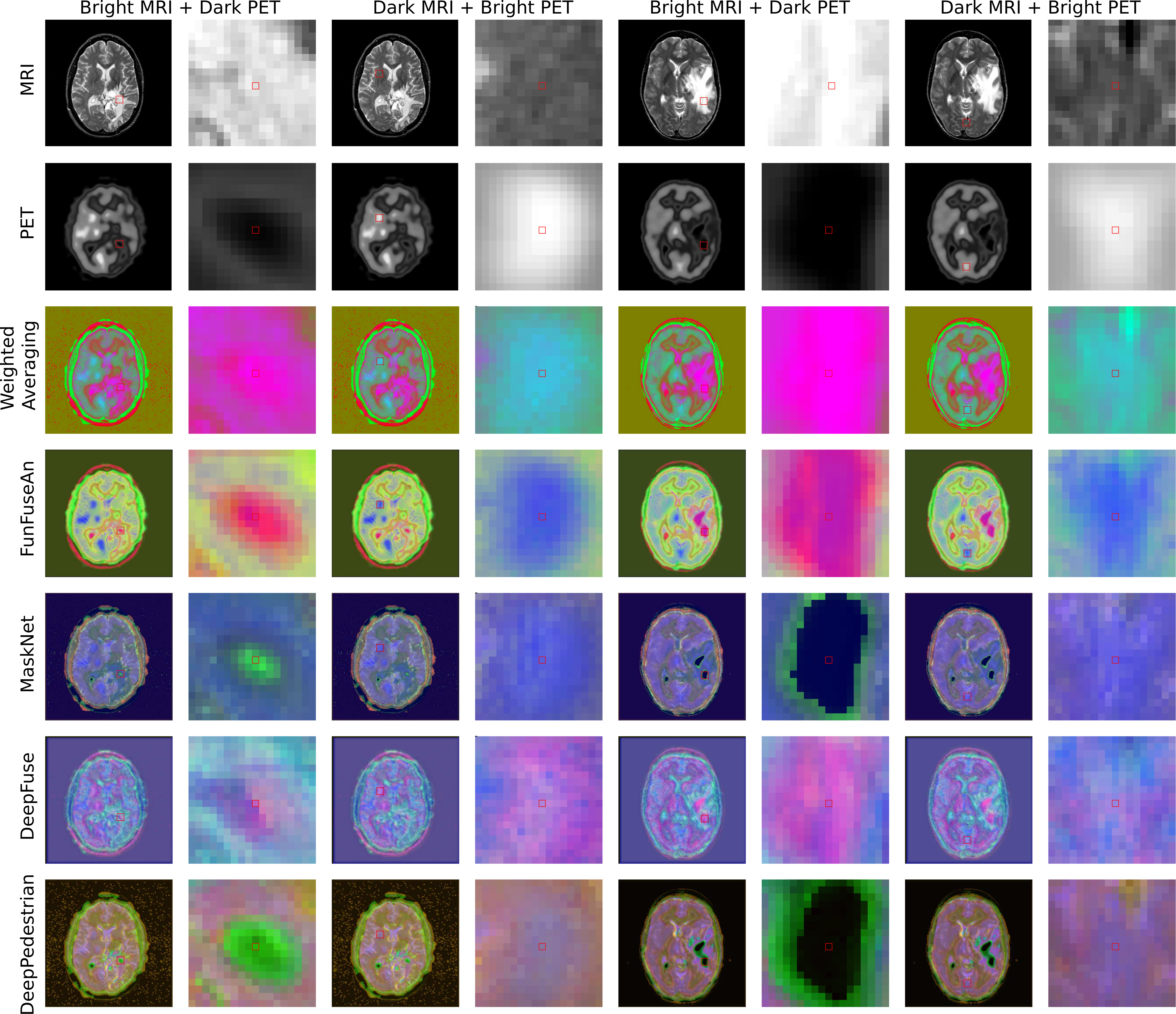}
\caption{The figure shows the guidance RGB images for each of the fusion methods.}
\label{fig:guidance RGB}
\end{figure}

In this section, we analyze the fusion methods with respect to its sensitivity to feature level changes in the input images by visualizing the \textit{Guidance MRI} and \textit{Guidance PET} images in Figure \ref{fig:guidance} and \textit{Guidance RGB} images in Figure \ref{fig:guidance RGB} for the two clinically relevant regions. We chose the first region as bright MRI with very dark PET features since it is the region that contains the necrotic core of glioma which clinicians are interested to operate for resection. The second region of interest has dark MRI with bright PET features resembling healthy tissues which is interesting for the visualization of high brain activity due to external stimuli. Ideally, a guidance MRI image should have low gradients meaning a fusion method should be less sensitive to changes in both dark and bright MRI regions.

\textit{Weighted Averaging}: For the region with bright MRI and dark PET intensities in both the test examples, it can be observed that the guidance MRI image is significantly brighter than the guidance PET image, revealing higher sensitivity to changes in MRI features. This suggests that a small change in the MRI pixel intensities will sharply modify the fused pixel intensities while it requires a substantial alteration in the PET pixel intensities to hold a considerable effect on the fused pixel intensities. This is an undesired outcome as a fusion method should be stable to changes in the pixel intensities of MRI. The guidance RGB image shows magenta color in these regions as the red channel containing guidance MRI image has high gradients and the blue channel containing the fused image have high pixel intensities compared to the green channel containing the guidance PET image with low gradients. On the other hand, the region with dark MRI and bright PET intensities shows an adverse effect where the guidance PET image is relatively brighter compared to the guidance MRI image since higher weightage is now given to the bright PET grayscale values. The guidance RGB image in this region shows cyan color since the green channel with guidance PET image has high gradients while the blue channel with the fused image also has higher intensities.

\textit{FunFuseAn}: For the region with bright MRI and dark PET intensities, there are higher gradients in the guidance MRI image compared to the guidance PET image. This conveys that the fused pixels within the necrotic core have a higher sensitivity to changes in the MRI pixels. The guidance RGB image also shows red color within the necrotic region indicating only the guidance MRI have high intensities compared to the images in other color channels. For the region with dark MRI and bright PET intensities, the guidance MRI and guidance PET images have low gradients. This means that the fused pixels in this region are quite stable to changes in the MRI and PET pixel intensities. Hence, the color in the guidance RGB image is predominantly blue as the fused pixel intensities are higher than the gradient values in both the guidance images. Therefore, it is challenging to interpret which of the two input modalities has a higher influence on the fused image in this particular region. 

\textit{DeepFuse}: In the bright MRI and dark PET region resembling necrotic core, there are relatively high gradient values in the guidance MRI image compared to the guidance PET image. This is the reason for the shades of magenta in the guidance RGB image. For the dark MRI and bright PET intensity region in the first test example, the pattern is similar where there are higher gradients in the guidance MRI image compared to the guidance PET image due to which magenta color is visible in the guidance RGB image. For the second test example; however, the guidance images from both MRI and PET have low gradients indicating a low influence of change in input pixel intensities within this region. Hence, the guidance RGB image also appears to be light blue in color in this region.

\textit{MaskNet \& DeepPedestrian}: For the necrotic region with bright MRI and dark PET intensities in the first test example, the guidance PET image is relatively brighter than the guidance MRI image for both \textit{MaskNet} and \textit{DeepPedestrian} networks. This conveys that both the networks are highly sensitive to the changes in very dark PET intensities of the necrotic region compared to the changes in MRI intensities. Since both the guidance MRI and the fused image have low intensities compared to the guidance PET image in this region, the guidance RGB image exhibits green color. However, for the second test example, both the guidance images have very low gradients in the same region, showcasing the stability of the networks for pixel intensity changes in this region. Since all the color channels in the guidance RGB image have low intensities, the guidance RGB image does not have a fixed color scheme and is dark.
For the region with dark MRI and bright PET intensities; however, the behavior is similar in both the test examples where both the guidance images have low gradients due to which it is challenging to differentiate the influence of each of the modalities while the \textit{MaskNet} network has learned the bright PET features better leading to blue color in the guidance RGB image.

\textit{Summary}: The guidance images provided a static overview of the influence of input principle pixel on the fused principle pixel and assisted in visualizing which of the two modalities has higher influence in the clinical region of interest. Therefore, it provided new insights related to the stability of fusion networks in these regions which were not perceivable by looking at the fused images. The guidance images clearly show that \textit{MaskNet} and \textit{DeepPedestrian} networks performed very differently compared to other fusion methods. For the region with the dark PET features resembling necrotic core, all the methods except \textit{MaskNet} and \textit{DeepPedestrian} were sensitive to changes in the bright MRI intensities which is not suitable for a reliable analysis of the necrotic tumor boundary since the dark PET features might not be properly preserved in the fused image due to changes in the MRI features. However, both \textit{MaskNet} and \textit{DeepPedestrian} preserved the dark PET features in the fused image and were quite stable to the changes in the MRI pixel intensities due to which both these methods are far more suitable for the clinical application than any other fusion methods. But the guidance images don't reveal the influence of neighborhood pixels on the fused principle pixel which is important for estimating a better fusion approach between the two of these networks.

\begin{figure}[!htb]
\centering
\includegraphics[width=15cm]{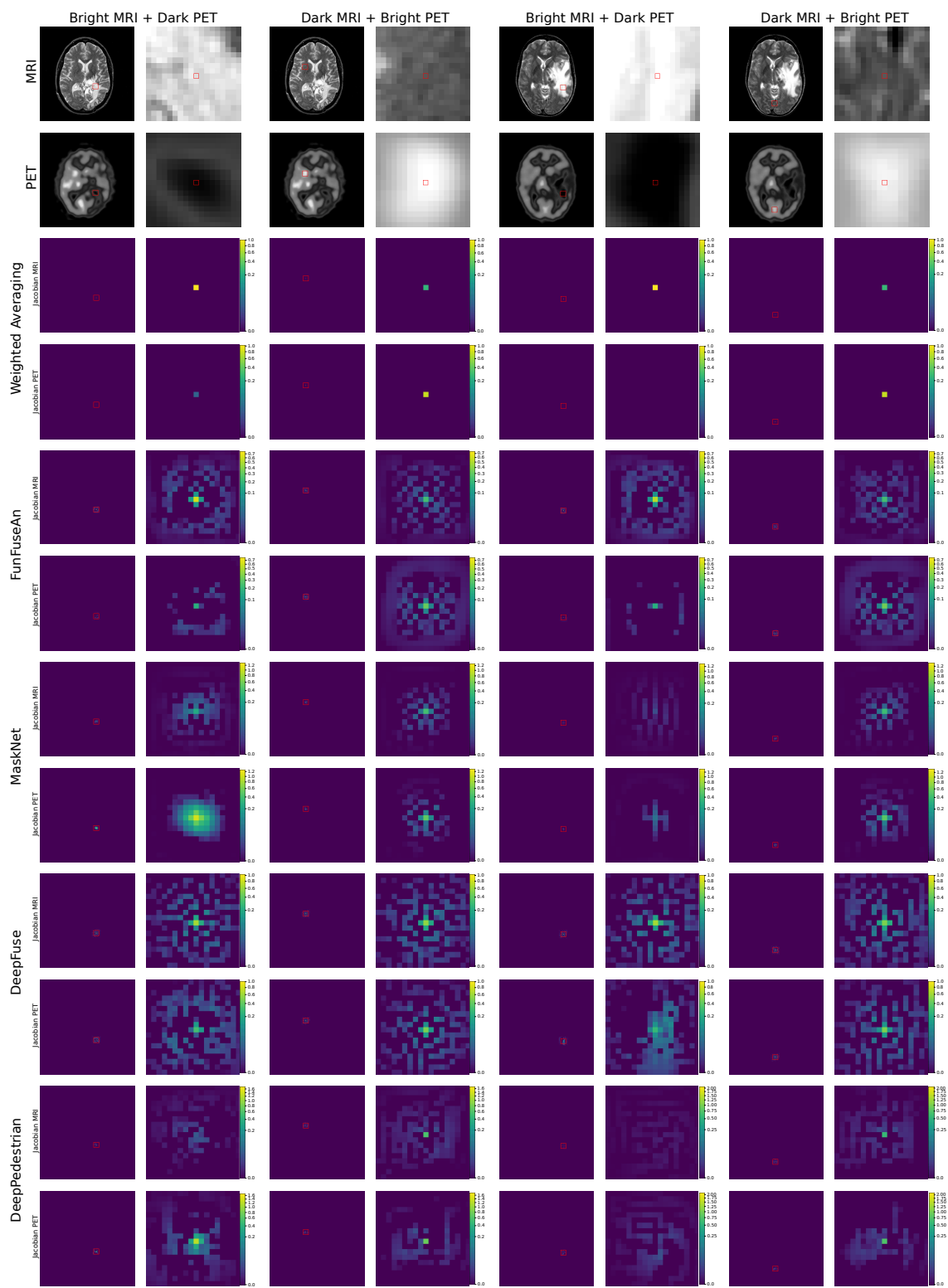}
\caption{The figure shows the jacobian MRI and jacobian PET images for each of the fusion methods in the clinical region of interests. The $\gamma_{corr1}$ was fixed at 0.3 for all the jacobian images.}
\label{fig:jacobian}
\end{figure}

\subsection{Jacobian images}
The jacobian images of the selected principle pixel is shown in Figure \ref{fig:jacobian}. The analysis of the jacobian images will help in understanding the influence of the neighborhood pixels located around the principle pixel in the clinically relevant regions of the input image. It will also assist in visually comparing the neighborhood influence of MRI pixels with PET pixels.

\textit{Weighted Averaging}: The jacobian MRI and jacobian PET images within the region with bright MRI and dark PET intensities for both the test examples show a higher gradient of the principle pixel in the jacobian MRI image compared to the jacobian PET image. For the dark MRI and bright PET intensities, there is a relatively high gradient in the principle pixel of the jacobian PET image showcasing sensitivity to changes in the bright PET pixel. One crucial observation from each of the jacobian images is that there is no influence of the neighborhood pixels on the prediction of the fused principle pixel since \textit{Weighted Averaging} performs the per-pixel computation of fused pixel intensities.  

\textit{FunFuseAn}: After analyzing the jacobian MRI and jacobian PET images within the region of bright MRI and dark PET intensities, it can be observed that similar to the \textit{Weighted Averaging} method, there is a higher gradient value for the principle pixel of the jacobian MRI image. The immediate neighborhood pixels in the jacobian MRI image also have significant gradient values meaning these pixels also influence the outcome of fused principle pixel. This is clinically useful since the change in intensities of pixels located within the necrotic core should affect the fused principle pixel. Additionally, there are more positive gradient values for the outer neighborhood pixels in the jacobian MRI image than the jacobian PET image. 
For the combination of dark MRI and bright PET intensities; however, there are low gradients for the principle pixel in both jacobian MRI and jacobian PET images. This conveys that the fused principle pixel is relatively stable to small changes in the bright PET and dark MRI pixel intensities. Hence, visualizing principle pixel in the jacobian images does not give much information about which input principle pixel has a higher influence on the fused principle pixel. On the other hand, the neighborhood pixel influence on the fused principle pixel is quite widespread due to the large number of positive gradients in both the jacobian MRI and jacobian PET images.

\textit{MaskNet}: For the combination of bright MRI and dark PET intensities, the jacobian images in the first test example convey quite distinct characteristics compared to the jacobian images in the second test example. For the first test example, the jacobian PET image has a much higher gradient for the principle pixel compared to the jacobian MRI image. This reveals a high sensitivity of the fused principle pixel within the necrotic tumor core with respect to small changes in the pixel intensity of the PET principle pixel whereas the fused principle pixel is quite stable to changes in the MRI principle pixel. Additionally, the influence of neighborhood pixels of PET within the dark necrotic tumor core on the fused principle pixel is highly intense compared to neighborhood pixels of MRI. This conveys that the changes in the pixel intensities of the PET inside the dark necrotic core can greatly affect the fused principle pixel while the PET pixels outside the necrotic core boundary have a negligible affect on the fused principle pixel. However, in the second test example, both the jacobian MRI and jacobian PET images have low gradients for the principle pixel as well as in the neighborhood region within the necrotic core. This demonstrates that the network requires quite significant changes in these features to change the fusion output. 
For the combination of dark MRI and high PET intensities in both the test examples, the jacobian MRI and jacobian PET have similar properties like \textit{FunFuseAn} where there are low gradients for the principle pixel in both the images; therefore, equally influencing the fused principle pixel. However, there are less number of neighborhood pixels in the jacobian MRI and jacobian PET images that influences the principle pixel when compared to the \textit{FunFuseAn} network. 

\begin{figure}[!htb]
\centering
\includegraphics[width=\textwidth]{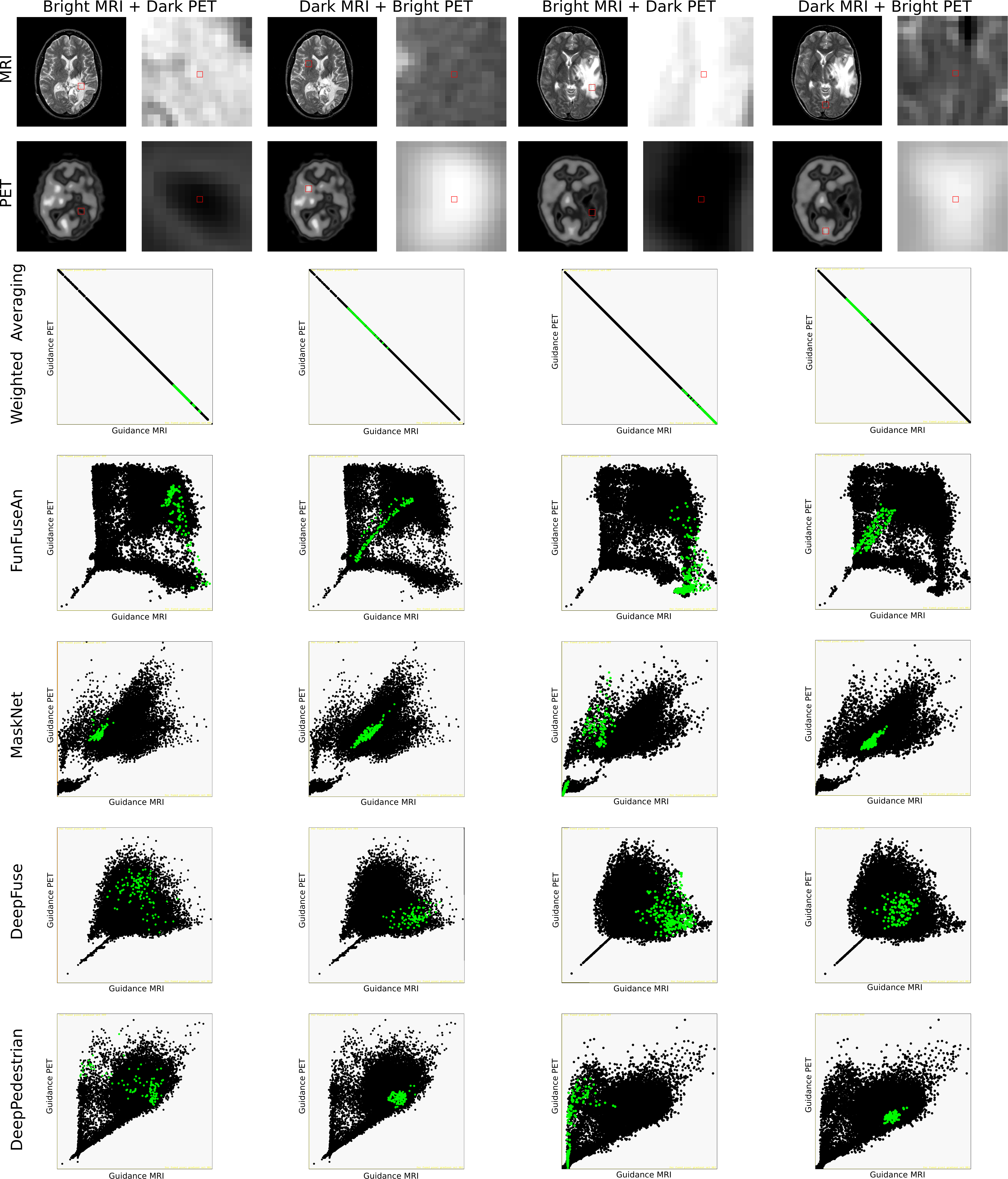}
\caption{The figure shows the scatterplots between the gradients of the guidance MRI and guidance PET images for each of the fusion methods. The green scatter points are the gradients for the pixels in the zoomed region of interest.}
\label{fig:scatterplots}
\end{figure}

\textit{DeepFuse}: The jacobian images in both the test examples, when visualized in the regions of bright MRI and dark PET intensities show a significantly high number of neighborhood pixels influencing the fused principle pixel while a high gradient value for the principle pixel can be seen in the jacobian MRI image compared to the jacobian PET image. Therefore, the fused principle pixel is highly sensitive to changes in the MRI pixel while it is stable to changes in the PET pixel. By analyzing the jacobian images for the combination of dark MRI and bright PET intensities, it is observable that the gradient values at the principle pixel are low; therefore, showcasing stability to the pixel intensity changes from both the modalities while widespread neighborhood pixels are influencing the fused principle pixel.

\textit{DeepPedestrian}: The combination of bright MRI and dark PET intensities in the first test example shows different jacobian results compared to the second test example. It can be noticed that the gradient value of the principle pixel in the jacobian MRI image is low while the gradient for the principle pixel is high in the jacobian PET image. This suggests that the fusion network is very stable to changes in the MRI pixel while sensitive to changes in the PET pixel within the necrotic tumor core. There are also positive gradients for the neighborhood pixels within the clinically important necrotic sub-region of glioma in the jacobian PET images due to which these pixels also influence the fusion principle pixel prediction. In the jacobian PET image, there is some influence from the neighborhood pixels outside the necrotic tumor core but they have low gradients compared to the gradients of the neighborhood pixels within the necrotic core. The jacobian images for the bright MRI and dark PET intensities in the second test example have very unique characteristics since both the jacobian MRI and jacobian PET images have very low gradients for the principle pixel as well as for the neighborhood pixels. Therefore, the network is very stable to changes in pixel intensities in this specific region of interest. For the combination of dark MRI and bright PET intensities, both the test examples show similar results where the neighborhood pixel influence on the fused principle pixel are relatively low in both the input modalities while the gradients for the principle pixel are also low meaning that the network does not get influenced by changes in pixels of each of the input modalities.

\textit{Summary}: The analysis of jacobian images helped to visually compare the influence of the input neighborhood pixels from each of the input modalities on the fused principle pixel. It was observed that most of the fusion based neural networks have a significant influence on neighborhood pixels. This visual analysis was not possible by only interpreting the guidance and fused images due to which jacobian images provide additional interpretation towards the suitability of a particular fusion method in a clinical setup. It was observed that \textit{Weighted Averaging} has no neighborhood influence which is understandable by the fact that it is a per-pixel computation scheme. The \textit{MaskNet} network; however, provides clinically favorable results as it has a high influence of neighborhood pixels within the necrotic core and very low influence from the pixels outside the necrotic core boundary.

\subsection{Scatterplots}
The scatterplots shown in Figure \ref{fig:scatterplots} helps in understanding the relationship between the gradients of guidance MRI and guidance PET images where each point resembles a pixel with the gradient value in guidance MRI at the horizontal axis and the gradient value in guidance PET at the vertical axis. 
A positive correlation between these gradients would mean that both the input modalities influence the fused pixel with equal strength. A negative correlation will show that an increase in the influence of one modality will lead to a decrease in the influence of the other modality. For the cases, where the gradients from one of the modalities are constant and only the gradient from the other modality varies, then this conveys that the feature in the input modality where the gradient varies is localized only in a subset of the pixels. If there is no correlation we cannot draw any conclusion from the scatterplot.

By examining the scatterplots for each of the fusion methods, no clear relationship between the guidance MRI and guidance PET images can be observed, except for the positive correlation between the low gradient values of guidance MRI and guidance PET images in the \textit{FunFuseAn} and \textit{DeepFuse} networks as well as a negative correlation between the gradients in the \textit{Weighted Averaging} method. However, by looking at the gradients within the zoomed region of interests shown by green points, a correlation patterns for few of the fusion networks can be determined. The scatterplot for \textit{Weighted Averaging} shows a negative correlation between the guidance MRI and guidance PET images. Therefore, for bright MRI and dark PET intensity setups in each of the two test examples, higher gradient values in guidance MRI causes the green scatter points to tilt towards the higher end of the horizontal axis while for dark MRI and bright PET intensity setup in the two test examples, the scatter points tilts in the opposite direction. This behavior shows that an increase in the influence of one modality led to a decrease in the influence of the other and \textit{Weighted Averaging} only prefers brightness as a feature which is clinically not interesting. The scatterplot of \textit{FunFuseAn} shows a positive correlation between the lower gradient values of guidance MRI and guidance PET and there is no correlation between the other gradient values of guidance MRI and guidance PET images. However, by visualizing the green points for dark MRI and bright PET intensities, we can observe a positive correlation between the gradient values which means both input modalities equally influence the fused pixel in this region. For the bright MRI and dark PET intensity setup, the scatter points are distributed in such a way that gradients from guidance MRI are high and constant while the gradients from the guidance PET varies.

The scatterplots for the \textit{MaskNet} network in bright MRI and dark PET intensity setup of the first test example show a positive correlation between the gradient values. In the second test example though, a positive correlation is seen only between the very low gradient values of guidance MRI and guidance PET images, while for higher gradients it is difficult to interpret a correlation. For the dark MRI and bright PET intensity setup, the gradients show a positive correlation in both the test examples.
The scatterplots for \textit{DeepFuse} does not show a clear correlation pattern between the gradients of guidance MRI and guidance PET images even after interpreting the green scatter points in the specific region of interests.  
For the second test example in the bright MRI and dark PET features of \textit{DeepPedestrian}, the gradients of guidance PET image varies comprehensibly while the gradients of guidance MRI image remains low and constant. In the dark MRI and bright PET intensity setup for both the test examples, \textit{DeepPedestrian} has gradients that are closely clustered and it is challenging to interpret a correlation.

\textit{Summary}: The scatterplots helped in interpreting the correlation between the gradients of the guidance MRI and guidance PET images for each of the fusion methods, which was difficult to estimate by visualizing only the jacobian and the guidance images. These correlations are important to visualize since an ideal fusion method should not have equal influence from both the input modalities in the regions where there are more clinically important features in one modality than the other. For example, in the region where there are dark PET features resembling the necrotic core, the gradients from guidance MRI should always be low and the scatterplot should not have a positive correlation between the gradients. The scatterplots of \textit{MaskNet} and \textit{DeepPedestrian} showed such characteristics but only for one of the test examples.

\subsection{Memory and Frame rates}
Table \ref{tab:latency} shows the timing results of the jacobian computations as well as the frame rates of our FuseVis tool for the visualization of jacobian images. As it can be seen from the results, the computation of jacobians is quite fast due to the powerful GPU hardware used during the experiments. The timings are averaged over 100 randomly selected principle pixels. The results unveil that \textit{FunFuseAn} and \textit{DeepFuse} are the fastest among the fusion based networks. One reason is that these networks have far fewer parameters to be optimized. Networks such as \textit{MaskNet} and \textit{DeepPedestrian} have a high number of trained parameters and dense hidden layers due to which the automatic differentiation through backpropagation requires higher computational time. We also estimated frame rates within a static framework where we iteratively saved the jacobian images for each of the principle pixels without using our FuseVis tool. Then, we displayed these images in the FuseVis tool during the mouseover operation. For this, we measured a frame rate of almost 40 frames per second ($fps$) which is expected considering there were no jacobian computations with backpropagation heuristics involved for each change of principle pixel during the mouseover operation. However, the disadvantage of such a setup is that it takes around 3 $GB$ of memory to save jacobian images locally for each input modality at 100 dots per inch ($DPI$) resolution whereas our tool requires zero additional memory usage. 
\begin{table}[H]
\caption{The table shows the timing and frame rate results for each of the fusion based neural networks.} 
\centering
\begin{tabular}{|P{4.5cm}|P{1.5cm}|P{1.3cm}|P{1.3cm}|P{2cm}|}
\hline Setting & FunFuseAn & MaskNet & DeepFuse & DeepPedestrian \\
\hline
 Jacobian computations & 0.003 s  & 0.004 s & 0.003 s &  0.005 s \\
\hline
FuseVis - Jacobian images 	& 20 fps & 15 fps & 20 fps & 10 fps \\
\hline
Guidance images & 198 s  & 265 s & 195 s & 360 s \\
\hline
\end{tabular}
\label{tab:latency}
\end{table}


\section{Conclusions}
In this work, we presented a novel interactive approach to visually inspect fusion networks. For achieving this, we developed an easy-to-use $\textit{FuseVis}$ tool that enables the end-user to visualize jacobian images of the selected principle pixel in real-time while performing a mouseover interaction. The real-time generation of jacobian images using an interactive user interface showcases the influence of pixels in the input images on a fused pixel. It is an intuitive idea that helps to overcome the huge computational complexity of generating per-pixel jacobian images that involves several backpropagation heuristics. The FuseVis tool also enables the user to visualize guidance images which estimate the sensitivity of image fusion networks to the changes in input features. Therefore, the guidance images are a very useful technique to get a concise overview of the huge information content in the jacobian images by analyzing the influence of the input pixel to the fused pixel at the same location for all pixels. We currently train different fusion architectures in a similar environmental setting to foster a suitable comparison of black box fusion algorithms for a critical application in medical diagnosis. However, our first of its kind visualization tool can easily be used to interpret any neural network in its original experimental setup. By visually analyzing the neural networks for image fusion within a specific clinical application, it was found that only \textit{MaskNet} and \textit{DeepPedestrian} provided results that would be helpful to clinicians. However, these methods do not preserve all clinically relevant features efficiently due to which these methods cannot be used in a clinical setup yet. Additionally, it is not trivial to attain fusion groundtruth of an ideal fused image based on the specific clinical requirements. Therefore, there is a need to train fusion based neural networks by using annotations provided by medical artists.
 We expect that our work should enable further research in the field of visual explanation and interpretation of fusion based neural networks as well as neural networks for other image processing based applications. 
\vspace{6pt} 



\funding{This work was supported by the European Social Fund and the Free State of Saxony under NeuroFusion grant (project no. 100312752) and SePIA grant (project no. 100299506).}

\acknowledgments{Some of the computations were also performed on an HPC  Cluster  at  the  Center  for  Information  Services  and High Performance Computing (ZIH) at TU Dresden.}

\conflictsofinterest{The authors declare that there is no conflict of interest.}



\reftitle{References}

\newpage
\begin{figure}[!htb]
\centering
\includegraphics[width=15cm]{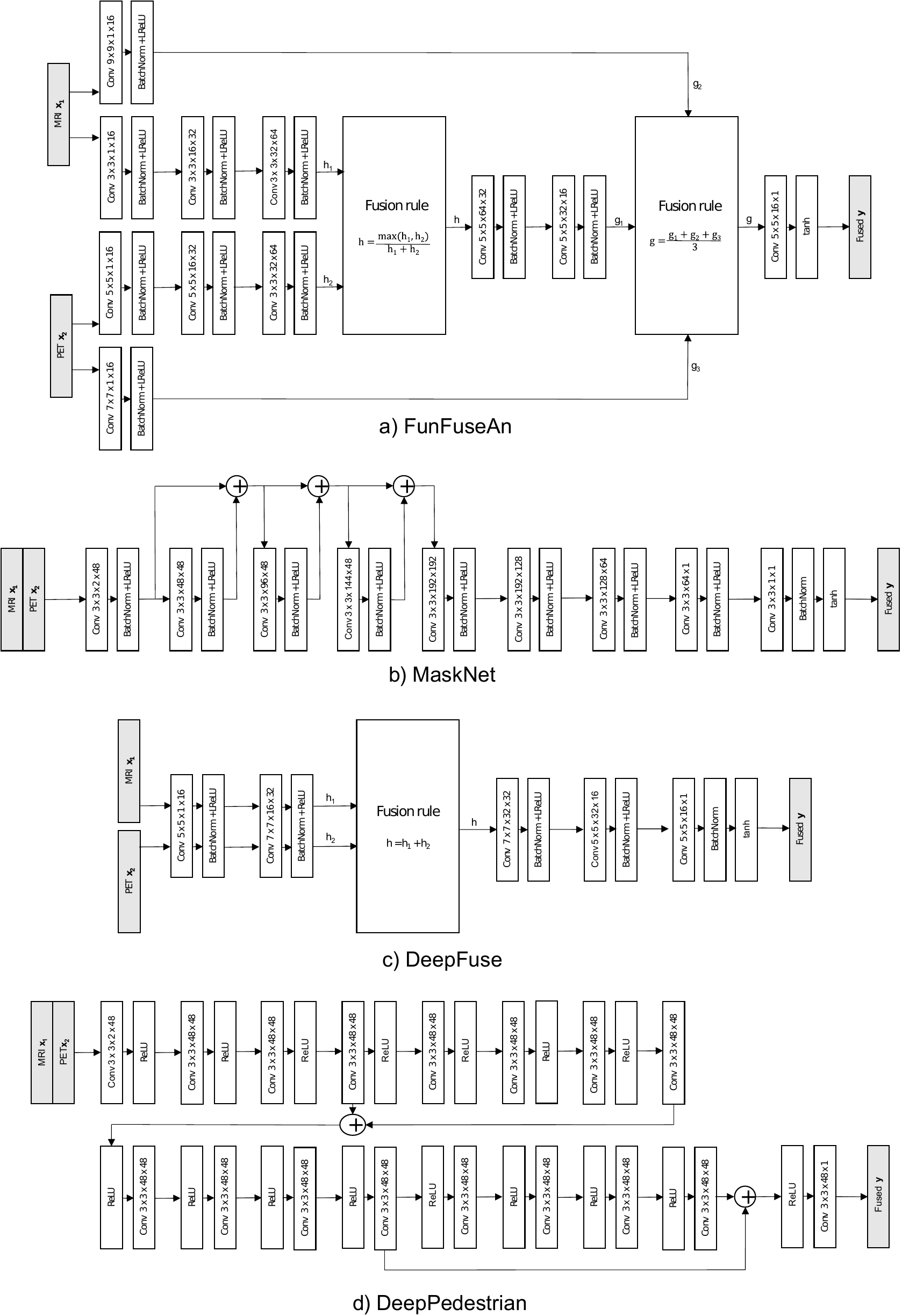}
\caption{The figure shows different neural network architectures used in this work.}
\label{fig:networks}
\end{figure}




\end{document}